\documentclass[10pt,twocolumn,letterpaper]{article}

\usepackage{cvpr}
\usepackage{times}
\usepackage{epsfig}
\usepackage{graphicx}
\usepackage{amsmath}
\usepackage{amssymb}
\usepackage{mathtools}
\usepackage{booktabs}
\usepackage[caption=false]{subfig}

\usepackage{mynotation}

\usepackage{algorithm}
\usepackage{algpseudocode}
\algrenewcommand{\algorithmicrequire}{\textbf{Input:}}
\algrenewcommand{\algorithmicensure}{\textbf{Output:}}

\usepackage[pagebackref=true,breaklinks=true,letterpaper=true,colorlinks,bookmarks=false]{hyperref}

\cvprfinalcopy % *** Uncomment this line for the final submission

\def\cvprPaperID{1494} % *** Enter the CVPR Paper ID here

\ifcvprfinal\fi
\begin{document}

%%%%%%%%% TITLE
\title{Adaptive Decontamination of the Training Set:\\A Unified Formulation for Discriminative Visual Tracking}

\author{Martin Danelljan, Gustav H\"ager, Fahad Shahbaz Khan, Michael Felsberg \\
	\small Computer Vision Laboratory, Department of Electrical Engineering, Link\"oping University, Sweden\\
	\small\{\texttt{martin.danelljan},\; \texttt{gustav.hager},\; \texttt{fahad.khan},\; \texttt{michael.felsberg}\}\texttt{@liu.se}
	{}
}

\maketitle
\thispagestyle{empty}

%%%%%%%%% ABSTRACT
\begin{abstract}
   Tracking-by-detection methods have demonstrated competitive performance in recent years. In these approaches, the tracking model heavily relies on the quality of the training set. 
Due to the limited amount of labeled training data, additional samples need to be extracted and labeled by the tracker itself. This often leads to the inclusion of corrupted training samples, due to occlusions, misalignments and other perturbations. Existing tracking-by-detection \mbox{methods} either ignore this problem, or employ a separate component for managing the training set. 

We propose a novel generic approach for alleviating the problem of corrupted training samples in tracking-by-detection frameworks. Our approach dynamically manages the training set by estimating the quality of the samples. Contrary to existing approaches, we propose a unified formulation by minimizing a single loss over both the target appearance model and the sample quality weights. The joint formulation enables corrupted samples to be down-weighted while increasing the impact of correct ones. Experiments are performed on three benchmarks: OTB-2015 with 100 videos, VOT-2015 with 60 videos, and Temple-Color with 128 videos. On the OTB-2015, our unified formulation significantly improves the baseline, with a gain of $3.8 \%$ in mean overlap precision. Finally, our method achieves state-of-the-art results on all three datasets. Code and supplementary material are available at \url{http://www.cvl.isy.liu.se/research/objrec/visualtracking/decontrack/index.html}.

\end{abstract}

\section{Introduction}
\label{sec:introduction}

Generic visual tracking is the problem of estimating the trajectory of a target in an image sequence, given only its initial location. 
Tracking methods serve as important components in a variety of vision systems.
The problem is particularly challenging due to the limited prior knowledge about the target. Furthermore, the tracking model must be flexible to counter rapid target appearance changes, while being robust to, \eg, occlusions and background clutter.

\begin{figure}%
	\centering\vspace{-5mm}
	\includegraphics*[trim=13 0 0 28,width=0.9\columnwidth,height=5.6cm]{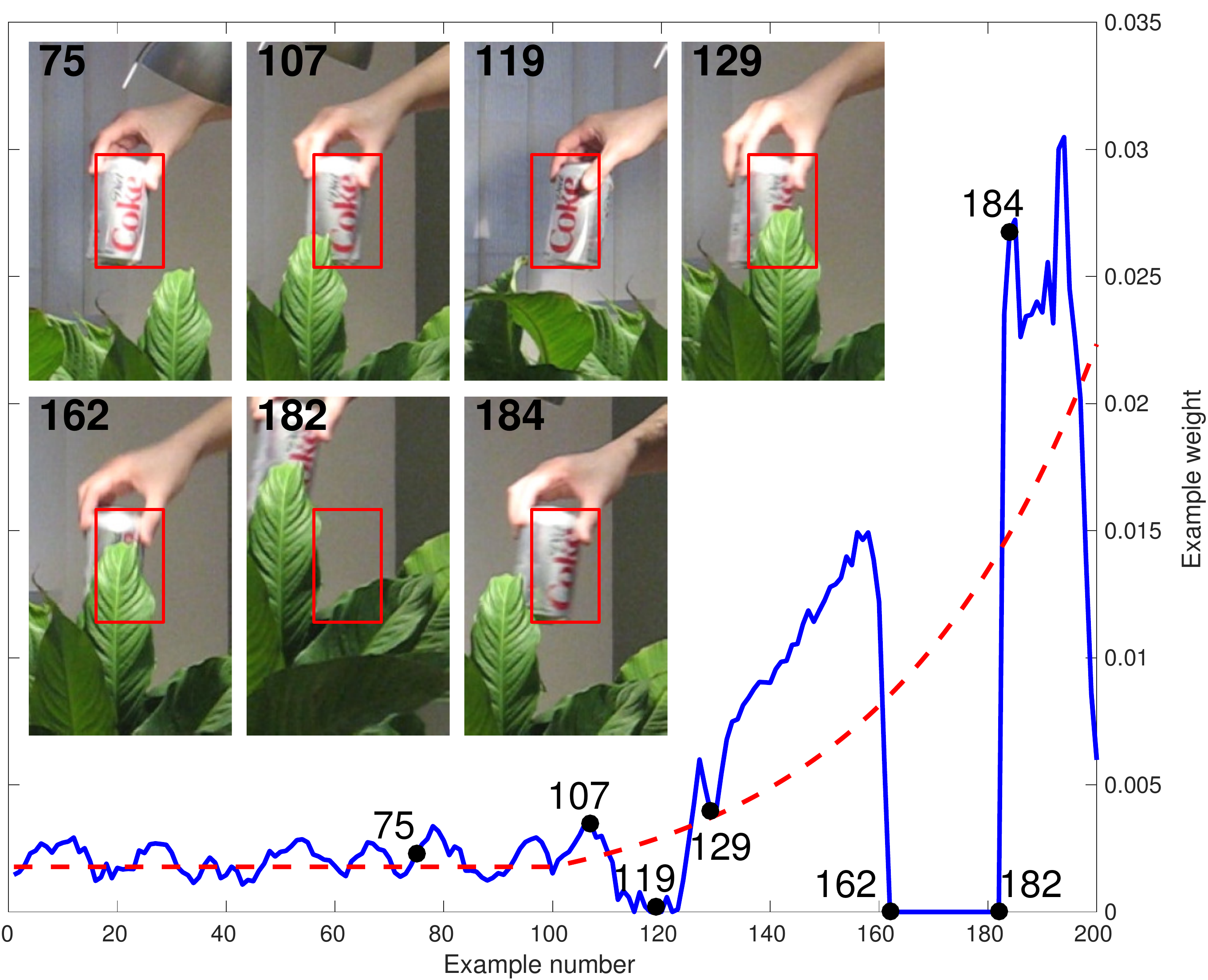}%
	\vspace{-1mm}
	\caption{An illustration of our adaptive decontamination of the training set. We show the corresponding image patches and tracking predictions (red box) for selected training samples. The quality weights (blue), computed by our learning approach, determine the impact of the corresponding training samples (numbered in chronological order). The prior sample weights are plotted in red. Our approach down-weights samples that are misaligned (no.\ 119), partially occluded (no.\ 129) or fully occluded (no. 162-182).}%
	\label{fig:intro_fig}
	\vspace{-4mm}
\end{figure}

The above mentioned problems have been addressed by methods based on the tracking-by-detection paradigm \cite{DanelljanBMVC14,Torr11b,Henriques14,MEEM2014}, with promising results in recent years. In this paradigm, tracking methods employ machine learning techniques to train an appearance model based on samples of the target and its background. Typically, supervised learning methods such as Support Vector Machines (SVMs) or ridge regression are used to construct a discriminative classifier or regressor. The quality of the tracking model is directly dependent on the training set. Therefore, a robust approach for constructing and managing the training set is crucial for avoiding model drift and tracking failure.

Standard tracking-by-detection approaches struggle when the training set is contaminated by corrupted samples. These corrupted samples are included in the training set in several different scenarios. Firstly, when encountered with target deformation and out-of-plane rotation, inaccurate tracking predictions lead to misaligned training samples (no.\ 119 in figure~\ref{fig:intro_fig}). Consequently, the model often drifts, eventually leading to tracking failure. Secondly, occlusions and clutter contaminate the positive training samples with background information, thereby deteriorating the discriminative power of the model (no.\ 162 in figure~\ref{fig:intro_fig}). In this work, we aim to enhance the robustness of standard tracking-by-detection approaches by tackling the problem of \emph{decontaminating} the training set. 

Existing discriminative trackers either ignore the problem of corrupted samples \cite{DanelljanBMVC14,Torr11b,TangICCV07} or employ an explicit training sample management component \cite{TGPR2014,MUSTer2015,Mikolajczyk10d,SelfPaced_CVPR13,MEEM2014}. A straightforward approach is to directly discard samples that do not meet a certain criterion \cite{MOSSE2010}. Other methods use a combination of experts \cite{Mikolajczyk10d,MEEM2014}, a separate tracking model \cite{MUSTer2015,LTC_CVPR15} or distance comparisons \cite{TGPR2014} for managing the training set. In this paper, we argue that the standard two-component strategy is suboptimal due to the reliance on heuristics. Instead, we revisit the standard tracking-by-detection formulation with the aim of integrating the estimation of the sample quality weights in the learning. 

\subsection{Contributions}
We propose a novel formulation for jointly learning the tracking model and the training sample weights. Our formulation is generic, and can be integrated into common supervised learning methods.
In each frame, a joint loss is minimized to update both the model parameters \emph{and} the importance weights. Our joint learning approach down-weights corrupted samples while increasing the importance of correct ones, as visualized in figure~\ref{fig:intro_fig}. Different from previous tracking methods, our unified formulation eradicates the need of an explicit sample management component.

To validate our approach, we perform extensive experiments on three benchmarks: OTB-2015 \cite{OTB2015} with 100 videos, VOT-2015 \cite{VOT2015} with 60 videos, and Temple-Color \cite{TempleColor} with 128 videos. Our unified approach demonstrates a significant gain of $3.8\%$ in mean overlap precision on OTB-2015, compared to the baseline. Further, our tracker achieves state-of-the-art results on all three datasets.

\section{Discriminative Tracking Methods}
In recent years, discriminative tracking-by-detection approaches \cite{DanelljanBMVC14,Torr11b,Henriques14,SelfPaced_CVPR13,MEEM2014} have shown promising results on benchmarks, such as OTB \cite{Wu13} and VOT \cite{VOT2014}. The appearance model within a tracking-by-detection framework is typically based on a discriminatively trained regressor \cite{DanelljanBMVC14,Torr11b,Henriques14} or classifier \cite{SelfPaced_CVPR13,MEEM2014}. These approaches are formulated in a supervised learning setting, where labeled training samples are collected from the sequence. Given a set of $n$ training examples $\{(x_j,y_j)\}_{j=1}^n$, the aim is to find the parameters $\theta \in \Omega$ of the appearance model. Here, $x_j \in \mathcal{X}$ denotes a feature vector in the sample space $\mathcal{X}$ and $y_j \in \mathcal{Y}$ is the corresponding label in the label set $\mathcal{Y}$. Many supervised learning methods in tracking \cite{DanelljanBMVC14,Torr11b,Henriques14,MEEM2014} find the parameter values $\theta$ by minimizing a loss of the form,
\begin{equation}
	\label{eq:supervised_learning}
	J(\theta) = \sum_{k=1}^{n} L(\theta; x_j, y_j) + \lambda R(\theta) .
\end{equation}
Here, $L : \Omega \times \mathcal{X} \times \mathcal{Y} \rightarrow \reals$ specifies the loss $L(\theta; x_j, y_j)$ for a training sample $(x_j,y_j)$ as a function of the parameters $\theta$. The impact of the regularization function $R : \Omega \rightarrow \reals$ is controlled by the constant weight $\lambda \geq 0$.

Eq.~\eqref{eq:supervised_learning} covers a variety of learning approaches, including support vector machines (SVMs) \cite{Torr11b,SelfPaced_CVPR13,MEEM2014} and discriminative correlation filters (DCFs) \cite{DanelljanSCIA2015,DanelljanVOT2015,DanelljanICCV2015,Henriques14}.
A common approach \cite{Mikolajczyk10d,SelfPaced_CVPR13,MEEM2014} is to use a two-class learning strategy to differentiate between the target $y_j = 1$ and background $y_j = -1$. Alternatively, the DCF based trackers \cite{DanelljanICCV2015,Henriques14}, utilize continuous labels $y_j \in [0, 1]$ or let $y_j$ be the desired confidence map over an image region. Another strategy \cite{Torr11b} is to let $\mathcal{Y}$ be the possible transformations of the target box.

\subsection{Training Sample Weighting}
\label{sec:sample_weighting}

In discriminative tracking, the model is learned using training samples collected from the video frames.
Typically, the training set is updated  with new samples in each frame, to account for changes in the target and background appearance. 
We rewrite \eqref{eq:supervised_learning} to highlight this temporal sampling used in many tracking methods. Let $(x_{jk},y_{jk})$ denote the $j$th training sample in frame number $k$. Assume that $n_k$ samples from frame $k \in \{1, \ldots, t\}$ are included in the training set, where $t$ denotes the current frame number. Typically, both positive and negative samples $(x_{jk},y_{jk})$ are extracted in a frame $k$, based on the estimated target location. The loss \eqref{eq:supervised_learning} is then expressed in the more general form,
\begin{equation}
	\label{eq:tracking_learning}
	J(\theta) = \sum_{k=1}^{t} \alpha_k \sum_{j=1}^{n_k} L(\theta; x_{jk}, y_{jk}) + \lambda R(\theta) .
\end{equation}
Here, the constant weights $\alpha_k \geq 0$ control the impact of samples from frame $k$. By increasing $\alpha_k$, a greater impact is given to samples $\{(x_{jk},y_{jk})\}_{j=1}^{n_k}$ extracted from frame $k$.

There exist several strategies to control the impact of training samples in \eqref{eq:tracking_learning}. In DCF-based trackers \cite{DanelljanBMVC14,Henriques14}, a learning rate parameter $\gamma \in [0, 1]$ is employed to update the weights as $\alpha_{k} = (1 - \gamma) \alpha_{k+1}$. Such a re-weighting strategy aims to reduce the impact of older samples in the learning. 
Trackers based on SVMs typically prune the training set by \eg rejecting samples older than a threshold \cite{TangICCV07} or removing support vectors with the least impact \cite{Torr11b}. However, these methods do not account for the problem of corrupted samples $(x_{jk},y_{jk})$ in the training set.

\subsection{Corrupted Training Samples} 
Contrary to object detection, the problem of corrupted training samples is commonly encountered in tracking. The problem appears since the samples are not hand-labeled, but rather labeled by the tracking algorithm itself. Several factors contribute to the unintentional inclusion of corrupted training samples in the learning.
(a) Inaccurate tracking predictions, due to \eg target rotations or deformations, lead to misaligned samples. This can result in model drift or tracking failure. (b) Partial or full occlusions of the target lead to positive samples being corrupted by the occluding objects. This is a common source of tracking failure, since the appearance model is contaminated due to occlusions. (c) Perturbations, such as motion blur, can lead to a distorted view of the target. These factors contribute to the inclusion of corrupted training samples in the learning, thereby deteriorating the discriminative power of the model.

\vspace{0mm}\noindent\textbf{State-of-the-Art:} Several recent works have investigated the problem of corrupted training samples in the tracking-by-detection paradigm \cite{MOSSE2010,Mikolajczyk10d,SelfPaced_CVPR13,MEEM2014}. Bolme \etal \cite{MOSSE2010} propose to reject new samples based on the Peak-to-Sidelobe Ratio (PSR) criterion. PSR is computed as the ratio between the maximum confidence score and the standard deviation of the surrounding scores (outside a specified neighborhood of the peak). Zhang \etal \cite{MEEM2014} use an entropy-based minimization to determine the best model in an expert ensemble. The ensemble consists of the current tracking model and snapshots from earlier frames. If a disagreement occurs, the expert with the minimum entropy criterion is selected as the new tracker model. Kalal \etal \cite{Mikolajczyk10d} tackle the drift problem by generating positive and negative samples based on spatial and temporal constraints. Supan\v{c}i\v{c} and Ramanan \cite{SelfPaced_CVPR13} propose a strategy for updating the training set by revisiting previously rejected samples. Hong \etal \cite{MUSTer2015} use a key-point based long-term memory component to detect occlusions and refresh the short-term memory. 

\vspace{0mm}\noindent\textbf{Differences to Our Approach:} As discussed above, existing tracking-by-detection approaches tackle the problem of corrupted samples with a dedicated separate component. This component is either based on distance comparisons  \cite{TGPR2014}, heuristics \cite{MOSSE2010,Zhong12g}, a set of experts \cite{Mikolajczyk10d,MEEM2014}, a separate tracking model \cite{MUSTer2015}, or model fitting \cite{SelfPaced_CVPR13}. Our approach differs from the aforementioned methods in several aspects. To the best of our knowledge, we are the first to propose a learning formulation that jointly optimizes the model parameters and the sample weights. Instead of binary decisions \cite{MOSSE2010,Mikolajczyk10d,SelfPaced_CVPR13,MEEM2014}, our approach is based on continuous importance weights. This enables us to down-weight the impact of corrupted training samples, while up-weighting correct ones. Further, our method allows mistakes to be corrected by redetermining the sample weights at each frame.

\section{Our Approach}
\label{sec:atsl}
Here, we propose our formulation for jointly learn the appearance model and the training sample weights in a tracking-by-detection framework.

\subsection{Motivation}
\label{sec:motivation}
To motivate our approach, we first distinguish three desirable characteristics to be considered when designing a method for decontaminating the training set.

\noindent\textbf{Continuous weights:} Most existing discriminative trackers \cite{MOSSE2010,Mikolajczyk10d,SelfPaced_CVPR13,MEEM2014} rely on binary decisions for including or removing potential training samples. 
This is problematic in ambiguous scenarios, such as moderate occlusions or slight misalignments (see figure~\ref{fig:intro_fig}), where the extracted samples are not entirely corrupted and still contain valuable information. Instead, continuous quality weights are expected to more accurately capture the importance of such samples.
\\ \noindent\textbf{Re-determination of Importance:} A common approach is to determine the importance of a sample based on previous frames only, \eg rejecting new samples based on the current appearance model \cite{MOSSE2010}. Ideally, all available information should be considered when updating the importance of a specific training sample, including more recent frames. By exploiting information from \emph{all} observed frames, the importance of older samples can be re-determined more accurately. This will enable previous mistakes to be corrected at a later stage in the tracking process. 
\\ \noindent\textbf{Dynamic Sample Prior:} Methods purely based on bottom-up statistics ignore prior knowledge associated with the samples. In cases of rapid target deformations and rotations, the tracker should emphasis recent samples for robustness. Dynamic prior knowledge is complementary to bottom-up information, and is expected to improve performance. 

\subsection{Problem Formulation}

Our approach jointly estimates both the model parameters $\theta$ and the weights $\alpha_k$. This is achieved by minimizing a single loss function $J(\theta, \alpha)$ to learn both the appearance model $\theta$ and the training sample weights $\alpha = (\alpha_1, \ldots, \alpha_t)$. To the best of our knowledge, we are the first to cast the problem of determining the sample quality in a joint optimization framework. We introduce the joint loss $J(\theta, \alpha)$,\vspace{-3mm}
\begin{subequations}
	\label{eq:atls}
	\begin{align}
	\label{eq:atls_loss}
	J(\theta, \alpha) = \sum_{k=1}^{t} & \alpha_k \sum_{j=1}^{n_k} L(\theta; x_{jk}, y_{jk}) + \frac{1}{\mu} \sum_{k=1}^{t} \frac{\alpha_k^2}{\rho_k} + \lambda R(\theta) \\
	\label{eq:weight_nonneg}
	\text{subject to} \hspace{5mm} & \alpha_k \geq 0 \,,\; k = 1, \ldots, t \\
	\label{eq:weight_sum}
	& \sum_{k=1}^{t} \alpha_k = 1 .
	\end{align}
\end{subequations}
Different from the standard weighted loss \eqref{eq:tracking_learning}, our formulation \eqref{eq:atls_loss} is a function of both the model parameters $\theta$ \emph{and} the sample weights $\alpha_k$. As a result, the weights $\alpha_k$ are no longer pre-determined constants. The constrains \eqref{eq:weight_nonneg} and \eqref{eq:weight_sum} ensure that the weights $\alpha_k$ are non-negative and sum up to one. The second term in the joint loss \eqref{eq:atls_loss} is a regularization term on the sample weights $\alpha$. This regularization is controlled by the flexibility parameter $\mu > 0$ and the prior sample weights $\rho_k > 0$, satisfying $\sum_k \rho_k = 1$. The parameter $\mu$ controls the adaptiveness of the example weights $\alpha$. Increasing $\mu$ leads to a higher degree of flexibility in the weights $\alpha$. We analyze the effect of $\mu$ and $\rho_k$ by considering the extreme cases of increasing ($\mu \rightarrow \infty$) and reducing ($\mu \rightarrow 0$) the flexibility parameter.

\vspace{1mm}\noindent\textbf{The case $\mu \rightarrow \infty$:} This corresponds to removing the second term in \eqref{eq:atls_loss}, implying no regularization on $\alpha$. For a fixed model parameter $\theta$, the loss \eqref{eq:atls} is then minimized by setting $\alpha_m = 1$ for the frame $m$ with the smallest total loss $\sum_{j=1}^{n_m} L(\theta; x_{jm}, y_{jm})$ and setting $\alpha_k = 0$ for $k \neq m$. The model will thus overfit to samples from a single frame $k = m$, if the second term in \eqref{eq:atls_loss} is removed. Therefore, it is imperative to use a regularization on the weights $\alpha$.

\vspace{1mm}\noindent\textbf{The case $\mu \rightarrow 0$:} By introducing Lagrange multipliers, it can be shown that $\alpha_k \rightarrow \rho_k$ when $\mu \rightarrow 0$, for a fixed $\theta$.\footnote{The proof is provided in the supplementary material.} Thus, reducing the parameter $\mu$ also reduces the flexibility of the weights $\alpha_k$ about the prior weights $\rho_k$. The standard weighted loss \eqref{eq:tracking_learning} is therefore obtained in the limit $\mu \rightarrow 0$ by setting $\alpha_k = \rho_k$. Our approach can be seen as a generalization of \eqref{eq:tracking_learning} by introducing flexible sample weights $\alpha_k$.

\subsection{Optimization}
\label{sec:optimization}

Here, we propose a strategy for solving the joint learning problem \eqref{eq:atls}. Our approach iteratively minimizes the loss by alternating between the model parameters $\theta$ and the example weights $\alpha$. This strategy is motivated by the fact that \eqref{eq:atls} is convex in the weights $\alpha$, given any fixed $\theta$. Further, many existing supervised learning approaches, such as SVM and DCF, rely on convex optimization problems \eqref{eq:supervised_learning}. It can be directly verified that \eqref{eq:atls} is biconvex if the weighted loss \eqref{eq:tracking_learning} is convex. That is, the optimization problem obtained by fixing either $\theta$ or $\alpha$ in \eqref{eq:atls} is convex. For biconvex problems, a standard approach is to use Alternate Convex Search (ACS) \cite{Biconvex}. In each frame, we perform $N$ ACS iterations to minimize our formulation \eqref{eq:atls}.
In each iteration, we solve the two convex subproblems obtained by fixing either $\alpha$ or $\theta$ in \eqref{eq:atls}. We call these steps ``Update $\theta$'' and ``Update $\alpha$''.

\vspace{1mm}\noindent\textbf{Update $\theta$:} We first describe the subproblem of finding the optimal $\theta$ given a fixed $\alpha = \alpha^{(i-1)}$. Here, $\alpha^{(i-1)}$ denotes the estimate of the weights $\alpha$ in iteration $i-1$ of the optimization. In the first iteration $i = 1$, the weights $\alpha^{(0)}$ are initialized using estimates from the previous frame. The subproblem obtained by fixing the weights $\alpha = \alpha^{(i-1)}$ in \eqref{eq:atls} corresponds to optimizing the weighted loss \eqref{eq:tracking_learning} with respect to $\theta$. This generates an updated model $\theta^{(i)}$. The optimization \eqref{eq:tracking_learning} is performed by the standard training method of the applied learning approach. 

\vspace{1mm}\noindent\textbf{Update $\alpha$:} The second step of iteration $i$ corresponds to optimizing \eqref{eq:atls} with respect to $\alpha$, while keeping $\theta = \theta^{(i)}$ fixed. By defining the total loss in frame $k$ by $L_{k}^{(i)} = \sum_{j=1}^{n_k} L(\theta^{(i)}; x_{jk}, y_{jk})$, the resulting subproblem is,
\begin{subequations}
	\label{eq:atls_alpha}
	\begin{align}
	\label{eq:atls_loss_alpha}
	\text{minimize} \hspace{5mm} &  J^{(i)}_2(\alpha) = \sum_{k=1}^{t} L_{k}^{(i)} \alpha_k   + \frac{1}{\mu} \sum_{k=1}^{t} \frac{\alpha_k^2}{\rho_k} \\
	\label{eq:weight_nonneg_alpha}
	\text{subject to} \hspace{5mm} & \alpha_k \geq 0 \,,\; k = 1, \ldots, t \\
	\label{eq:weight_sum_alpha}
	& \sum_{k=1}^{t} \alpha_k = 1 .
	\end{align}
\end{subequations}
The above optimization problem can be efficiently solved with convex quadratic programming methods. We use the corresponding function in Matlab's Optimization Toolbox, which internally employs the interior point method. 

\begin{algorithm}[t]
	\caption{Our approach: tracking in frame $t$}
	\begin{algorithmic}[1]
		\Require
		Current model parameters $\theta$ and weights $\{\alpha_k\}_{k=1}^{t-1}$.
		\Statex \hspace{4.5mm} Current training set $\{(x_{jk},y_{jk})\}_{j=1,k=1}^{n_k,t-1}$.\vspace{1mm}
		\State Estimate the target location in frame $t$ using $\theta$.
		\State Extract training samples $\{(x_{jt},y_{jt})\}_{j=1}^{n_t}$ in frame $t$.
		\State Update the prior weights $\{\rho_k\}_{k=1}^t$ using, \eg, \eqref{eq:prior_weights}.
		\State Initialize weights $\alpha_k^{(0)} = \alpha_k$ for $k<t$ and $\alpha_t^{(0)} = \rho_t$.
		\For{$i = 1, \ldots, N$}
		\State \textbf{Update $\theta$:} Find $\theta^{(i)}$ by optimizing \eqref{eq:tracking_learning} using $\alpha^{(i-1)}$.
		\State \textbf{Update $\alpha$:} Find $\alpha^{(i)}$ by solving \eqref{eq:atls_alpha} given $\theta^{(i)}$.
		\EndFor
	\end{algorithmic}
	\label{alg:tracking}
\end{algorithm}

\begin{figure*}[!t]
	\centering
	\includegraphics[width=0.49 \textwidth]{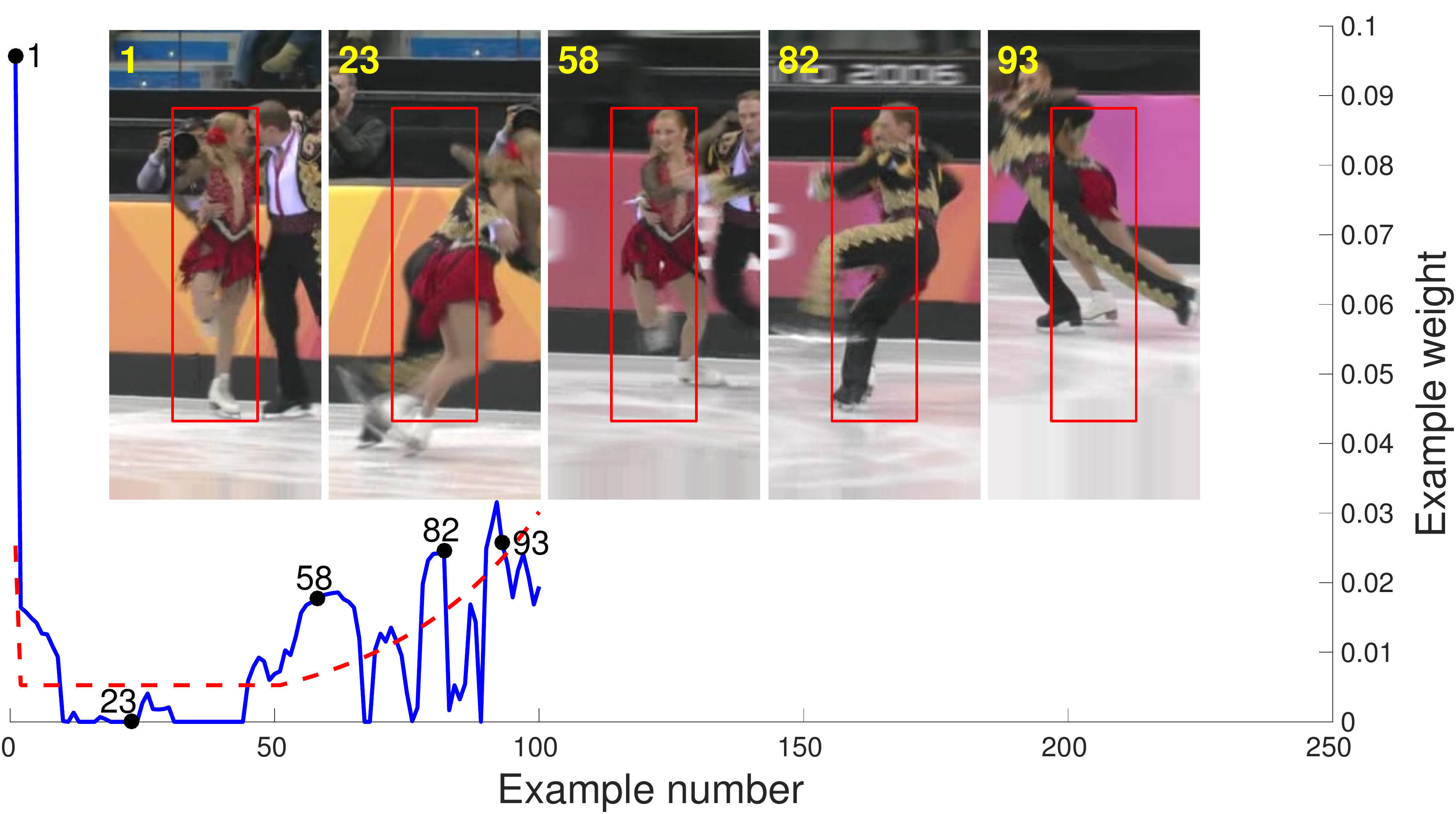}
	\includegraphics[width=0.49 \textwidth]{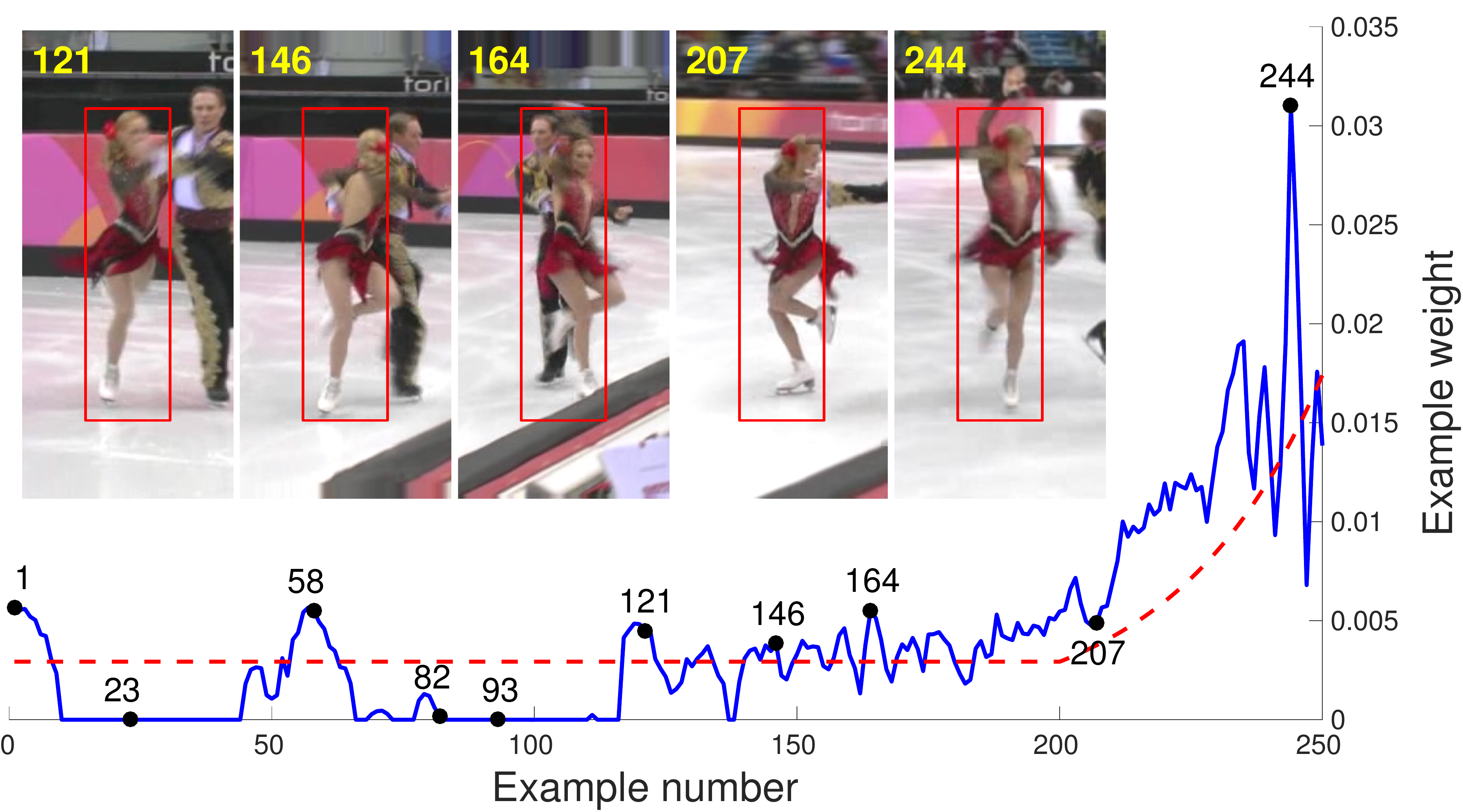}\vspace{-1mm}
	\caption{The training sample impact weights computed by our joint learning formulation on the \emph{Skating} sequence. The computed weights $\alpha_k$ (blue curve) and corresponding prior weights $\rho_k$ (red curve) are plotted for two different time instances during the tracking process: in frame 100 (left) and frame 250 (right). Image patches and corresponding target estimations (red box) are shown for example frames. The parameters are set as described in section~\ref{sec:details}. A few training examples (\eg no.\ 82 and 93) that are corrupted by an occluding object (the male skater) are initially assigned large weights (left). By efficiently redetermining all impact weights $\alpha_k$ in each frame, previous mistakes are corrected. In this example, the corrupted samples (no.\ 82 and 93) are down-weighted at a later stage (right). On the other hand, accurate training samples (no.\ 1 and 58) are consistently assigned large impact weights.}
	\label{fig:weights_method}
	\vspace{-3mm}
\end{figure*}

\subsection{Prior Weights Selection}
\label{sec:prior_weights}
As discussed in section~\ref{sec:motivation}, it is desirable to encode prior knowledge about the sample weights $\alpha_k$ in the learning. In our approach, this prior information is incorporated using the prior weights $\rho_k$, which serve as a regularizer for the sample weights $\alpha_k$.
The impact of the prior weights $\rho_k$ are further controlled by the flexibility parameter $\mu$.
We propose a simple, yet effective strategy for setting the sample weights $\rho_k$ based on solely temporal information. In our strategy, recent samples are given larger prior weights to account for fast appearance changes. In general, additional information about the sampling process, such as the number of training samples $n_k$ in frame $k$, can be integrated into $\rho_k$.

We use a learning rate parameter $\eta \in [0, 1]$ to determine the prior weights for the $K$ most recent frames, such that $\rho_{k} = (1 - \eta)\rho_{k+1}$ for $k = t - K, \ldots , t-1$. The prior weights for all frames older than $t - K$ are set to constant, \ie $\rho_{k} = \rho_{k+1}$ for $k < t - K$. The above recursive definition implies the formula
\begin{equation}
	\label{eq:prior_weights}
	\rho_k = 
	\begin{dcases}
		a \, , \; & k = 1, \ldots, t - K - 1 \\
		a (1 - \eta)^{t - K - k} \, , \; & k = t - K, \ldots, t .
	\end{dcases}
\end{equation}
Here, the constant $a = \left(t - K + \frac{(1 - \eta)^{-K} - 1}{\eta}\right)^{-1}$ is determined by the condition  $\sum_k \rho_k = 1$. The prior weights $\rho_k$ in \eqref{eq:prior_weights} put a larger emphasis on recent frames to alleviate the problem of rapid appearance changes, caused by \eg target deformations and out-of-plane rotations. Instead of letting the prior weights decrease towards zero for older samples, we assign all samples older than $K$ frames equal prior importance. This allows a significant influence of old training samples in the learning. Algorithm~\ref{alg:tracking} provides an overview of our method in a generic setting.

\section{The Tracking Framework}
Here, we describe a tracking-by-detection framework using the unified learning formulation proposed in section~\ref{sec:atsl}.

\subsection{Baseline Tracker}
\label{sec:baseline_tracker}
In recent years, the Discriminative Correlation Filter (DCF) based trackers have shown excellent performance on several benchmark datasets \cite{DanelljanBMVC14,DanelljanICCV2015,Henriques14,VOT2014}. These trackers apply Fourier transforms to efficiently train a discriminative regressor on sample feature maps extracted from the video frames. We employ the recent SRDCF \cite{DanelljanICCV2015} as our base supervised learning approach. Unlike other DCF methods, SRDCF employs a spatial regularization in the learning to alleviate the periodic effects induced by circular correlation.

The appearance model of the SRDCF tracker consists of a discriminatively trained convolution filter. In each new frame, a confidence map is first computed by applying the filter around the predicted target location. This confidence map is then maximized to estimate the target location. A single training example $(x_k,y_k)$ is added in each frame $k$. The sample $x_k$ is a $d$-dimensional feature map, extracted around the target, that also includes the surrounding background information. The Gaussian label function $y_k$ contains the desired confidence map, when applying the sought convolution filter $f_\theta$ on $x_k$. In the SRDCF, the model parameters $\theta$ thus consist of the filter coefficients in $f_\theta$. The standard SRDCF employs the weighted learning formulation \eqref{eq:tracking_learning}, with a per-sample loss $L$ given by,
\begin{equation}
	\label{eq:srdcf_example_loss}
	L(\theta; x_k, y_k) = \bigg\| y_k - \sum_{l=1}^{d} f_\theta^l \conv x_k^l \bigg\|^2 .
\end{equation}
Here, $\conv$ denotes circular convolution and the superscript $x_k^l$ and $f_\theta^l$ denotes the $l$th channel of $x_k$ and $f_\theta$ respectively. The loss \eqref{eq:srdcf_example_loss} consists of the total squared error between the desired confidence map $y_k$ and the confidence scores obtained by applying the filter $f_\theta$ to the sample $x_k$.

The regularization $R(\theta)$ is determined by the spatial penalty function $w$, consisting of a positive penalization factor at each spatial location in the filter,
\begin{equation}
	\label{eq:srdcf_regularization}
	R(\theta) = \sum_{l=1}^{d} \big\| w \pmult f_\theta^l \big\|^2 .
\end{equation}
Here, $\pmult$ denotes pointwise multiplication. The regularization ensures a limited spatial extent of the filter by penalizing coefficients outside the target region. The filter $f_\theta$ is trained by transforming the loss \eqref{eq:tracking_learning} to a real-valued Fourier basis and solving the resulting normal equations.
For more details about the SRDCF tracker, we refer to \cite{DanelljanICCV2015}.

\subsection{Proposed Tracker}
Here, we apply our unified learning formulation to the baseline tracker. To learn the appearance model, the baseline tracker minimizes the weighted loss \eqref{eq:tracking_learning}, using exponentially decaying sample weights $\alpha_k$. Instead, we minimize the proposed unified formulation \eqref{eq:atls} to jointly estimate both the model $\theta$ and the sample weights $\alpha_k$, in each frame. 

The proposed tracker follows the outline in algorithm~\ref{alg:tracking}. In a new frame $t$, we first estimate the target location as in the baseline tracking approach. The training set is then augmented with the new sample $(x_t, y_t)$. The prior weights $\rho_k$ are then updated as in \eqref{eq:prior_weights}. The importance weight of the new sample is initialized with its prior weight $\alpha_t^{(0)} = \rho_t$. The weights $\alpha_1^{(0)}, \ldots, \alpha_{t-1}^{(0)}$ of earlier samples are initialized with their estimates from the previous frame and then normalized such that $\sum_k \alpha_k^{(0)} = 1$. To minimize the joint loss \eqref{eq:atls}, we then perform an ``Update $\theta$'' step followed by an ``Update $\alpha$'' step in each iteration $i$ of the optimization, as described in section~\ref{sec:optimization}. As mentioned in section~\ref{sec:optimization}, our joint learning formulation \eqref{eq:atls} is biconvex since the weighted loss \eqref{eq:tracking_learning} is convex.

\vspace{0mm}\noindent\textbf{Update $\theta$:} The updated filter $f_\theta^{(i)}$ is computed using the training procedure in \cite{DanelljanICCV2015}, given the weights $\alpha_k^{(i-1)}$. Instead of the incremental update, we use the general formula in \cite{DanelljanICCV2015} to compute the normal equations. This enables arbitrary weights to be used. A fixed number of Gauss-Seidel iterations are then performed, with the current filter $f_\theta^{(i-1)}$ as an initial solution, to obtain the new filter $f_\theta^{(i)}$. 

\vspace{0mm}\noindent\textbf{Update $\alpha$:} The new filter $f_\theta^{(i)}$ is then used to redetermine sample weights $\alpha^{(i)}$. Since each frame only contains a single sample, the total loss in frame $k$ is given by $L_k^{(i)} = L(\theta^{(i)}; x_k, y_k)$. This is efficiently computed using the Fast Fourier Transform (FFT), by applying Parseval's formula to \eqref{eq:srdcf_example_loss}. The new weights $\alpha^{(i)}$ are then computed by solving the quadratic programming problem \eqref{eq:atls_alpha}.

To achieve an upper bound on the memory consumption, we store a maximum number $T$ of training samples. If the number of samples exceeds $T$, we simply remove the sample $k \leq K$ that has the smallest weight $\alpha_k$. Figure~\ref{fig:weights_method} shows the estimated quality weights $\alpha_k$ for an example sequence.

\section{Experiments}
\label{sec:experiments}
To evaluate our approach, we perform comprehensive experiments on three benchmark datasets: OTB-2015 \cite{OTB2015}, VOT-2015 \cite{VOT2015} and Temple-Color \cite{TempleColor}.\footlabel{exp}{Detailed results are presented in the supplementary material.}

\subsection{Parameter Settings}
\label{sec:details}
The prior sample weights are set as described in section~\ref{sec:prior_weights}, using $K = 50$ and $\eta = 0.035$. In general, the flexibility parameter $\mu$ depends on the scale of the loss function $L(\theta;x,y)$ for different discriminative methods. This dependency can however be mitigated by appropriate normalization of $L$ with, \eg, respect to the average number of samples $n_k$ per frame. We use $\mu = 5$ in our experiments, which enables a large degree of adaptiveness in the weights (see figure~\ref{fig:intro_fig} and \ref{fig:weights_method}). The maximum number of stored training samples is set to $T = 300$. In the tracking scenario, the joint loss \eqref{eq:atls} is modified marginally in each frame by adding the new training samples for frame $t$. Therefore, we found a single ($N=1$) ACS iteration to be sufficient to refine the estimate of $\theta$ and $\alpha$ from the previous frame. This further ensures a minimal increase in computations compared to the original learning approach.
Our joint learning is started at $t = 10$ frames into the sequence. This ensures a sufficient number training samples for our learning procedure.

For the baseline tracker \cite{DanelljanICCV2015}, we use the Matlab implementation provided by the authors. For a fair comparison, we use the same parameter settings for both our tracker and the compared baseline SRDCF. For the OTB-2015, we use HOG features for both our and the baseline tracker, as in \cite{DanelljanICCV2015}. For VOT-2015 and Temple-Color, we employ the same combination of HOG and Color Names for both trackers and use $\mu = 3$ and $T = 200$ in our method. Furthermore, we fix the parameter settings for all videos in each dataset. In our approach, solving the quadratic programming problem \eqref{eq:atls_alpha} in the ``Update $\alpha$'' step, is highly efficient. It takes around 5 milliseconds on a standard desktop computer. The computational cost of our tracker is completely dominated by the baseline training procedure used in the ``Update $\theta$'' step. We obtain a slightly reduced frame rate  (around 3 frames per second) compared to the baseline tracker.

\begin{table}
	\centering
	\resizebox{0.48\textwidth}{!}{%
		%\begin{tabular}{|l||c|c|c|c|c|}
%\hline
%&SRDCF&SRDCF prior&SRDCF MEEM&SRDCF PSR&\textbf{SRDCF ATSL}\\\hline\hline
%Mean OP&72.9&72.9&72.2&\textit{\textcolor{blue}{74.4}}&\textbf{\textcolor{red}{76.5}}\\\hline
%Mean DP&78.9&\textit{\textcolor{blue}{79.2}}&77.5&79.1&\textbf{\textcolor{red}{82.3}}\\\hline
%Mean CLE&\textit{\textcolor{blue}{38.6}}&37.8&\textbf{\textcolor{red}{44.4}}&35&29.8\\\hline
%Mean AUC&60.5&60.6&60.1&\textit{\textcolor{blue}{60.9}}&\textbf{\textcolor{red}{63.3}}\\\hline
%Mean FPS&4.29&1.52&\textit{\textcolor{blue}{5.23}}&\textbf{\textcolor{red}{12.4}}&2.31\\\hline
%\end{tabular}

\begin{tabular}{lcccc}
\toprule
&Baseline \cite{DanelljanICCV2015}&Baseline-Entropy&Baseline-PSR&\textbf{Ours}\\\midrule
Mean OP&72.9&72.2&74.4&\textbf{\textcolor{red}{76.7}}\\\bottomrule
%Mean DP&78.9&\textit{\textcolor{blue}{79.2}}&77.5&79.1&\textbf{\textcolor{red}{82.3}}\\\hline
%Mean CLE&\textit{\textcolor{blue}{38.6}}&37.8&\textbf{\textcolor{red}{44.4}}&35&29.8\\\hline
%Mean AUC&60.5&60.6&60.1&\textit{\textcolor{blue}{60.9}}&\textbf{\textcolor{red}{63.3}}\\\hline
%Mean FPS&4.29&1.52&\textit{\textcolor{blue}{5.23}}&\textbf{\textcolor{red}{12.4}}&2.31\\\hline
\end{tabular}
	}
	\vspace{0.5mm}
	\caption{A comparison of our approach, using mean OP ($\%$), with the baseline methods on the OTB-2015 dataset. The baseline tracker does not account for corrupted samples. We also compare our approach by incorporating the entropy and PSR strategies in the baseline tracker. The best result is displayed in red font. Our approach achieves a significant performance gain of $3.8\%$ in mean OP, compared to the baseline tracker.}
	\label{tab:baseline}
	\vspace{-2mm}
\end{table}

\begin{table*}[!t]
	\centering
	\resizebox{1.004\textwidth}{!}{%
		\begin{tabular}{l@{~}c@{~~}c@{~~}c@{~~}c@{~~}c@{~~}c@{~~}c@{~~}c@{~~}c@{~~}c@{~~}c@{~~}c@{~~}c@{~~}c@{~~}c@{~~}c@{~~}c@{~~}c}
\toprule
&EDFT\cite{Felsberg13c}&LSHT\cite{Shengfeng13b}&DFT\cite{Laura12d}&ASLA\cite{Jia12d}&TLD\cite{Mikolajczyk10d}&Struck\cite{Torr11b}&CFLB\cite{GaloogahiCVPR2015}&ACT\cite{DanelljanCVPR14}&TGPR\cite{TGPR2014}&KCF\cite{Henriques14}&DSST\cite{DanelljanBMVC14}&SAMF\cite{Li2014}&DAT\cite{possegger15a}&MEEM\cite{MEEM2014}&LCT\cite{LTC_CVPR15}&HCF\cite{HCF_ICCV15}&SRDCF\cite{DanelljanICCV2015}&\textbf{Ours}\\\midrule
OTB-2015&41.4&40.0&35.9&49.0&46.5&52.9&44.9&49.6&54.0&54.9&60.6&64.7&36.4&63.4&70.1&65.5&\textit{\textcolor{blue}{72.9}}&\textbf{\textcolor{red}{76.7}}\\
Temple-Color&41.2&29.0&34.0&38.8&38.4&40.9&37.8&42.1&51.6&46.5&47.5&56.1&48.2&\textit{\textcolor{blue}{62.2}}&52.8&58.2&\textit{\textcolor{blue}{62.2}}&\textbf{\textcolor{red}{65.8}}\\\bottomrule
\end{tabular}

	}
	\vspace{0mm}
	\caption{A comparison of our approach, using mean OP ($\%$), with state-of-the-art trackers on the OTB-2015 and Temple-Color datasets. The best two results are shown in red and blue font respectively. On the OTB-2015 dataset, the best existing tracker provides a mean OP of $72.9\%$. On the Temple-Color dataset, both SRDCF and MEEM obtains a mean OP score of $62.2\%$. Our approach obtains state-of-the-art results, outperforming the best existing trackers by $3.8\%$ and $3.6\%$, on the OTB-2015 and Temple-Color datasets, respectively.}
	\label{tab:OTB100}
	\vspace{-3mm}
\end{table*}

\subsection{Baseline Experiments}
We first compare our approach with the baseline SRDCF tracker, which does not account for corrupted training samples. We also integrate two existing training sample management strategies into the baseline tracker, for additional comparisons. The first strategy \cite{MOSSE2010} is based on the Peak to Sidelobe ratio (PSR) of the tracking confidence scores. It is computed as the ratio $\frac{g_\text{max}-m_r}{\sigma_r}$, where $g_\text{max}$ is the maximum confidence, $m_r$ is the mean confidence and $\sigma_r$ is the standard deviation of the confidence scores outside the peak.
The second strategy \cite{MEEM2014} is based on an expert ensemble of previous snapshots of the tracking model. In each frame, the confidence scores are first computed for all experts. If the target location estimates differ, an entropy based score is used to rank the experts in the ensemble. The current tracking state is then set to the expert with the highest score. This corresponds to resetting the tracker model to the best previous state. New snapshots are stored periodically, while discarding the oldest one. For a fair comparison, we optimize the parameters for the PSR and entropy based strategies.

We report the results using mean overlap precision (OP). The OP is computed as the percentage of frames where the intersection-over-union (IOU) overlap with the ground-truth exceeds a threshold of $0.5$. Table~\ref{tab:baseline} shows the mean OP results over all the 100 videos of OTB-2015 dataset. The baseline SRDCF tracker obtains a mean OP score of $72.9\%$. The PSR strategy improves the results with a mean OP score of $74.4\%$. Our approach significantly improves the performance by providing a gain of $3.8\%$ in mean OP, compared to the baseline tracker. The substantial improvement over the baseline demonstrates the importance of decontaminating the training sample set. It is worth to mention that our approach is generic, and can be incorporated into other discriminative tracking frameworks.

\begin{figure}[!t]
	\centering\vspace{-3.5mm}
	\newcommand{\wid}{0.24\textwidth}
	\subfloat[OTB-2015\label{fig:sota_otb}]{\includegraphics[width = \wid]{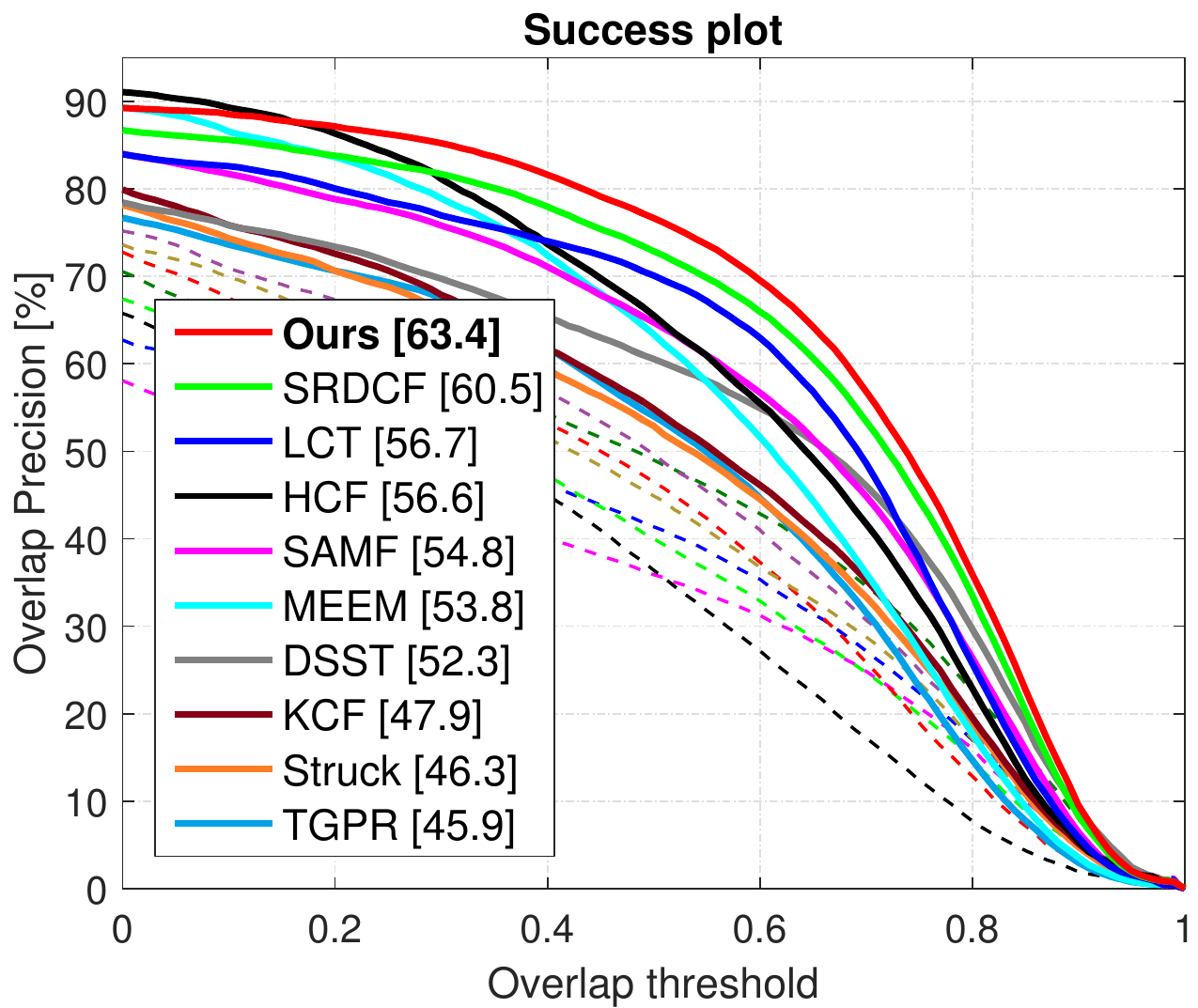}}
	\subfloat[Temple-Color\label{fig:sota_tpl}]{\includegraphics[width = \wid]{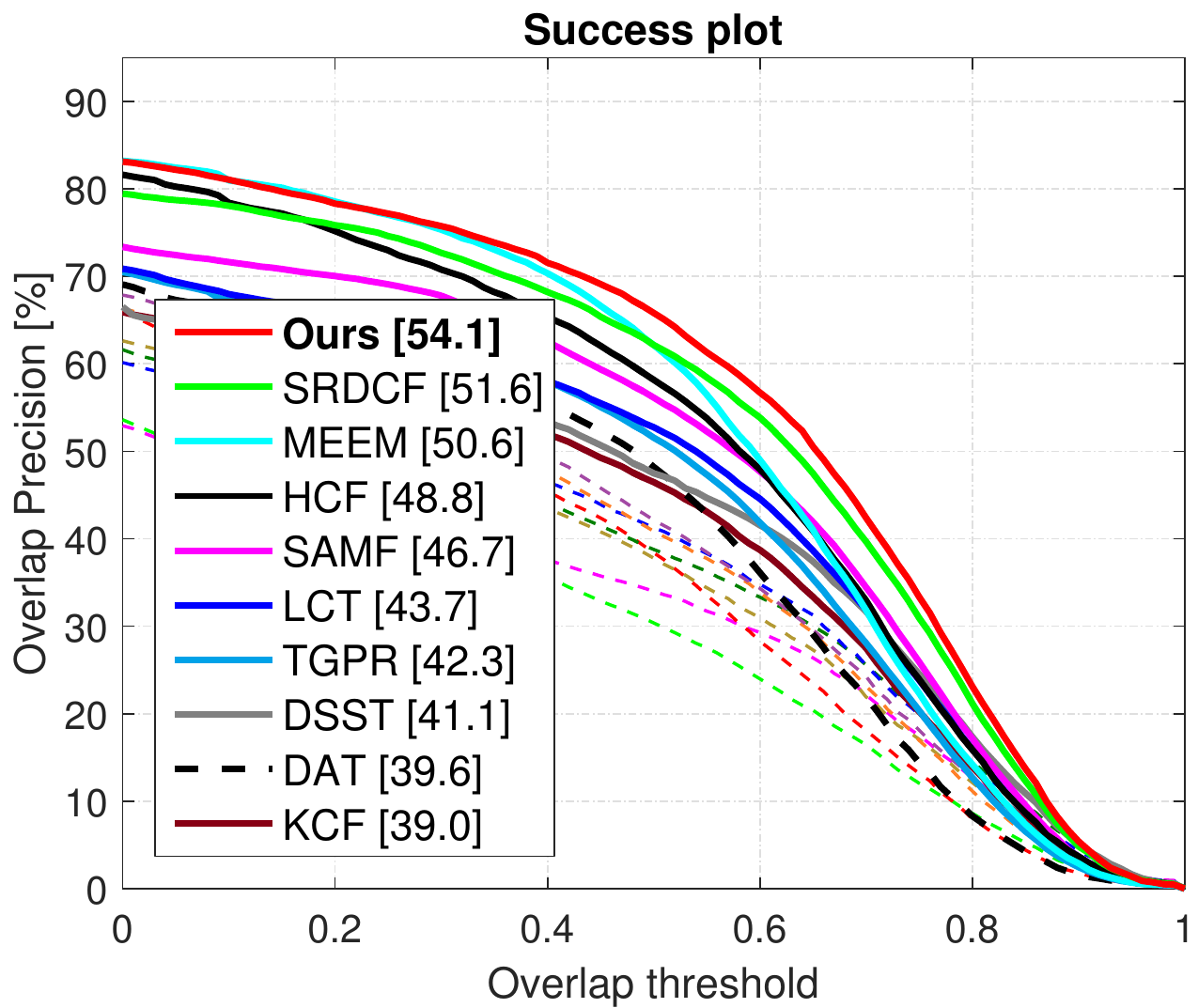}}\vspace{1mm}
	\caption{Success plots for the OTB-2015 (a) and Temple-Color (b) datasets. For clarity, we only show the top 10 trackers in the legend. On the OTB-2015 and Temple-Color, our approach achieves state-of-the-art results with a gain of $2.9\%$ and $2.5\%$ in AUC respectively, compared to the best previous method.}\vspace{-2mm}
	\label{fig:OTB_tpl}
\end{figure}

We also validate the generality of our approach by applying the proposed joint learning formulation to an SVM-based discriminative model. SVMs have been successfully applied for tracking-by-detection in recent years \cite{Torr11b,SelfPaced_CVPR13,MEEM2014}. We use a binary linear SVM, where $L(\theta;x,y)$ is the standard hinge-loss. As in the SRDCF case, we use the outline described in algorithm~\ref{alg:tracking} and set the prior weights $\rho_k$ as in \eqref{eq:prior_weights}. In each frame $k$, we collect 1 positive and about 20 negative samples $(x_{jk},y_{jk})$ from the estimated target neighborhood, using the color-based feature representation \cite{MEEM2014}. For the baseline SVM tracker, we fix the sample weights as $\alpha_k = \rho_k$. For our SVM tracker, we minimize the loss \eqref{eq:atls} as described in section~\ref{sec:optimization}. The same parameter settings is used for both the baseline and our version of the SVM-based tracker. On OTB-2015, the baseline SVM tracker provides a mean OP of $58.2\%$. Our SVM tracker achieves a significant gain of $3.2\%$, with a mean OP of $61.4\%$.

\subsection{OTB-2015 Dataset}
We perform a comprehensive comparison with 17 recent state-of-the-art trackers: EDFT \cite{Felsberg13c}, LSHT \cite{Shengfeng13b}, DFT \cite{Laura12d}, ASLA \cite{Jia12d}, TLD \cite{Mikolajczyk10d}, Struck \cite{Torr11b}, CFLB \cite{GaloogahiCVPR2015}, ACT \cite{DanelljanCVPR14}, TGPR \cite{TGPR2014}, KCF \cite{Henriques14}, DSST \cite{DanelljanBMVC14}, SAMF \cite{Li2014}, DAT \cite{possegger15a}, MEEM \cite{MEEM2014}, LCT \cite{LTC_CVPR15}, HCF \cite{HCF_ICCV15} and SRDCF \cite{DanelljanICCV2015}.

\subsubsection{State-of-the-art Comparison}

A comparison with state-of-the-art trackers on the OTB-2015 is shown in Table~\ref{tab:OTB100} (first row). We report the mean OP over all the 100 videos. The MEEM tracker, with an entropy minimization based sample management, obtains a mean OP of $63.4\%$. The hierarchical convolutional features (HCF) tracker provides a mean OP of $65.5\%$. Our approach significantly outperforms the best compared tracker, by achieving a mean OP of $76.7\%$.

\begin{figure}[!t]
	\centering
	\newcommand{\wid}{0.12\textwidth}
	\includegraphics*[trim = 0 100 40 0, width = \wid]{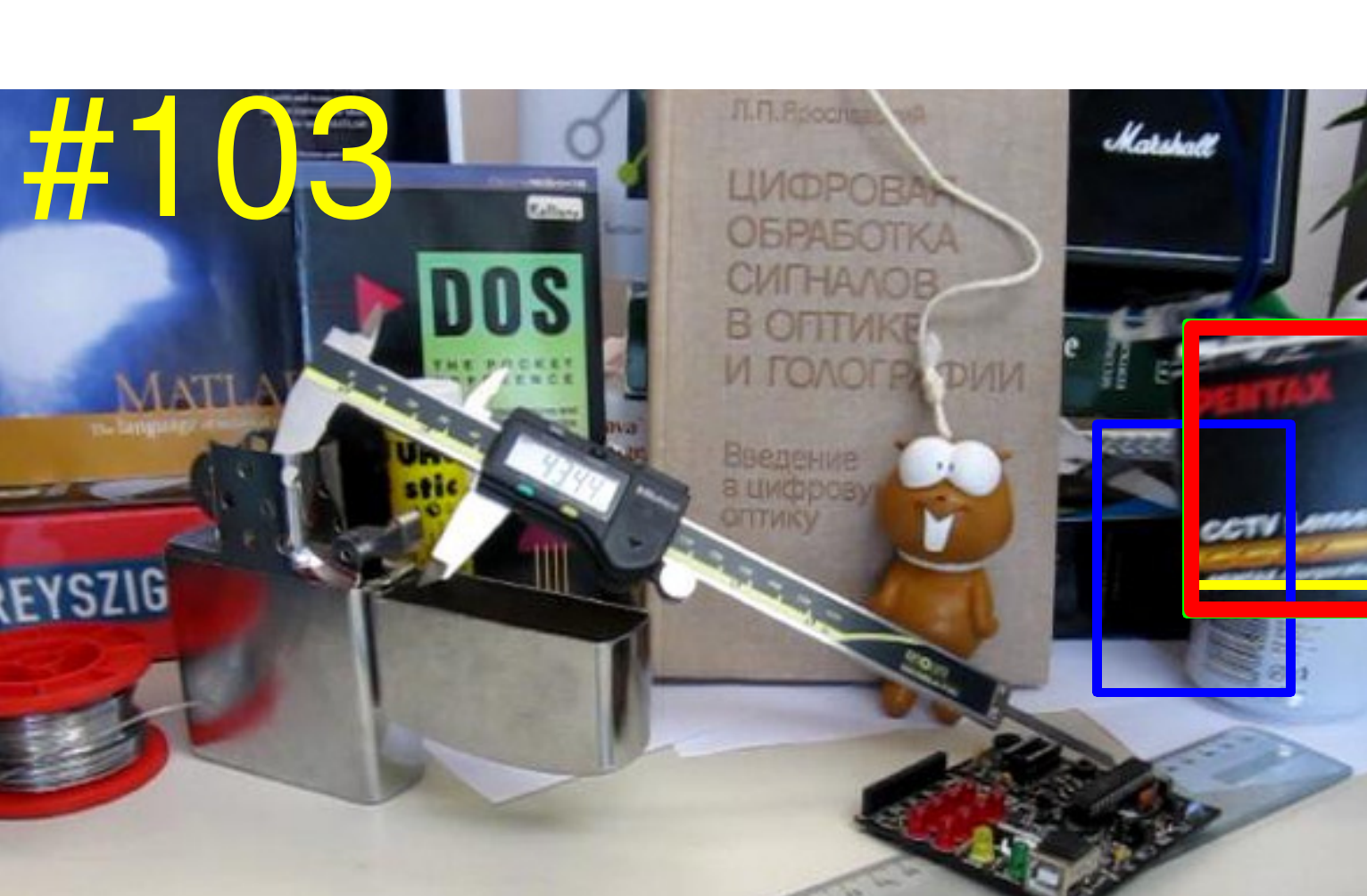}%
	\includegraphics*[trim = 0 100 40 0, width = \wid]{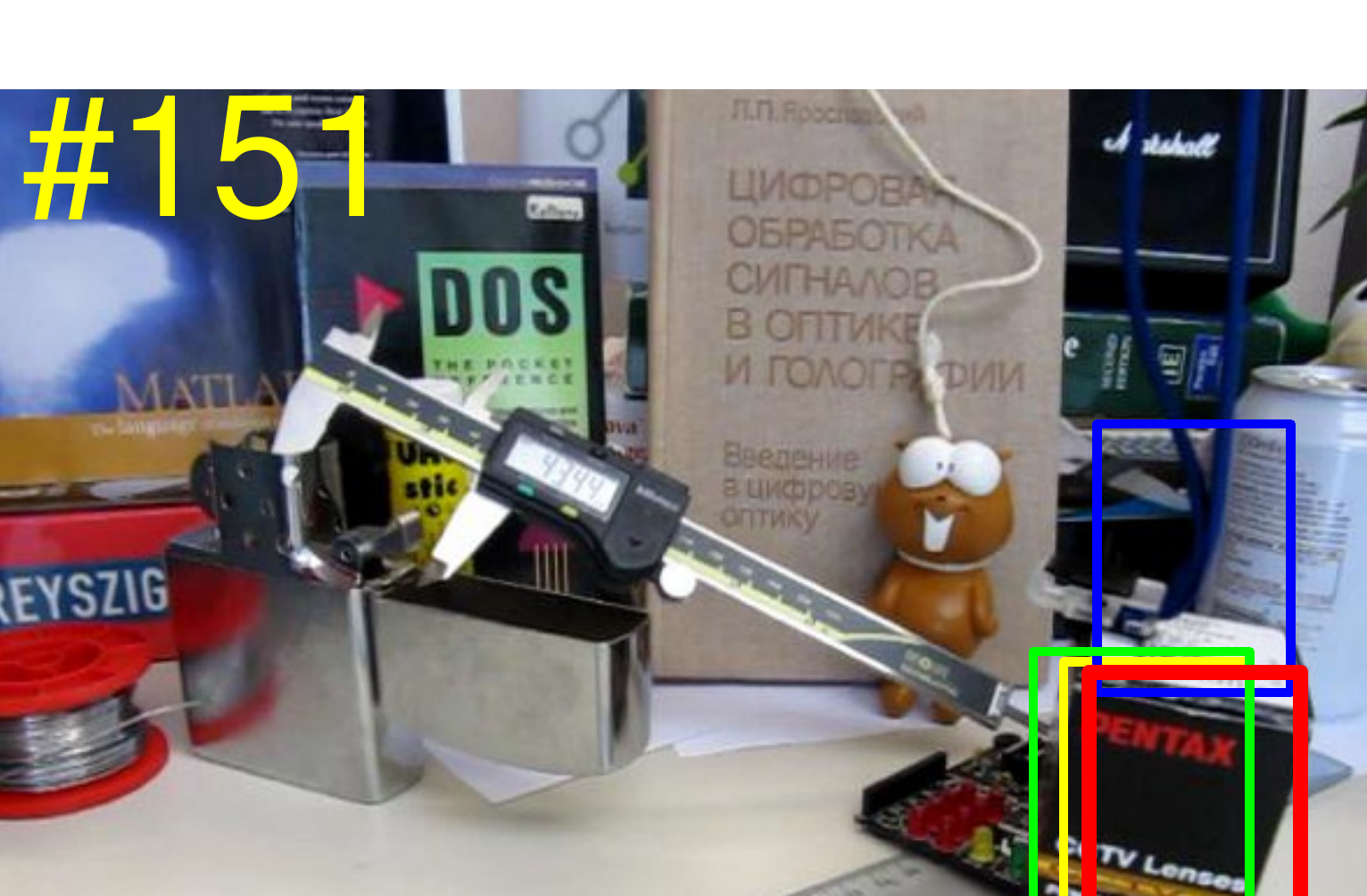}%
	\includegraphics*[trim = 0 100 40 0, width = \wid]{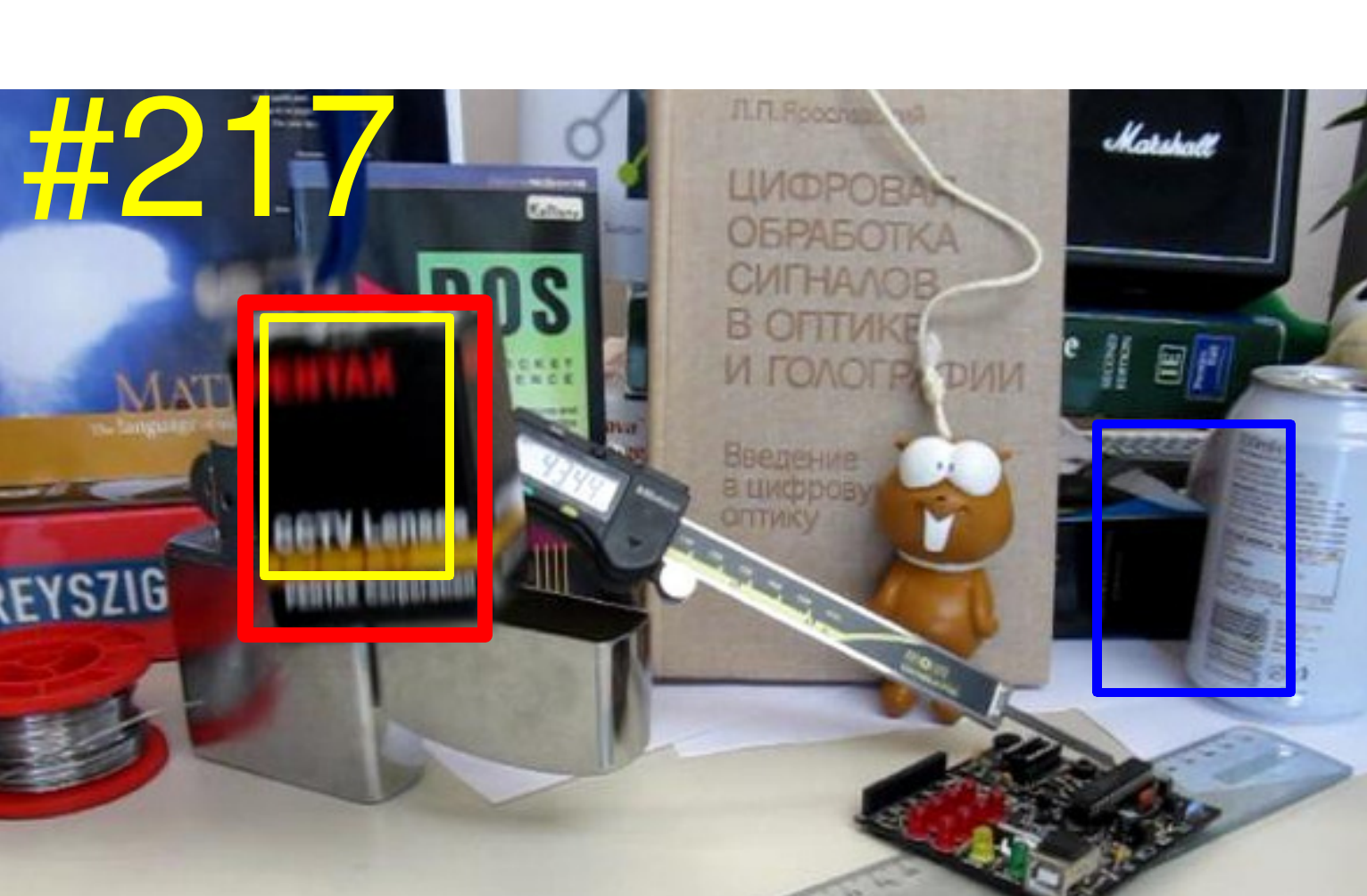}%
	\includegraphics*[trim = 0 100 40 0, width = \wid]{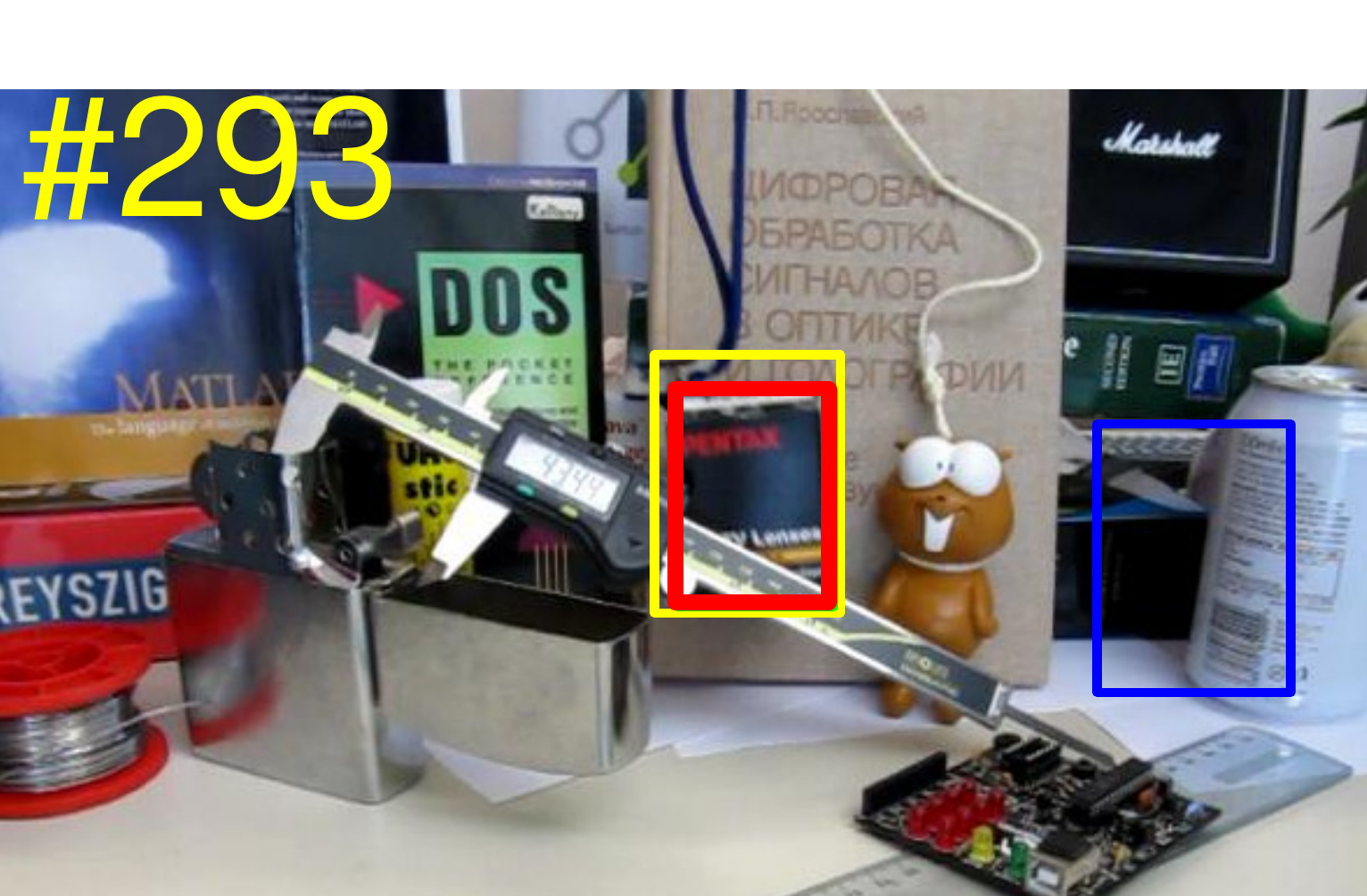}\vspace{-0.5mm}
	\includegraphics*[trim = 0 100 50 0, width = \wid]{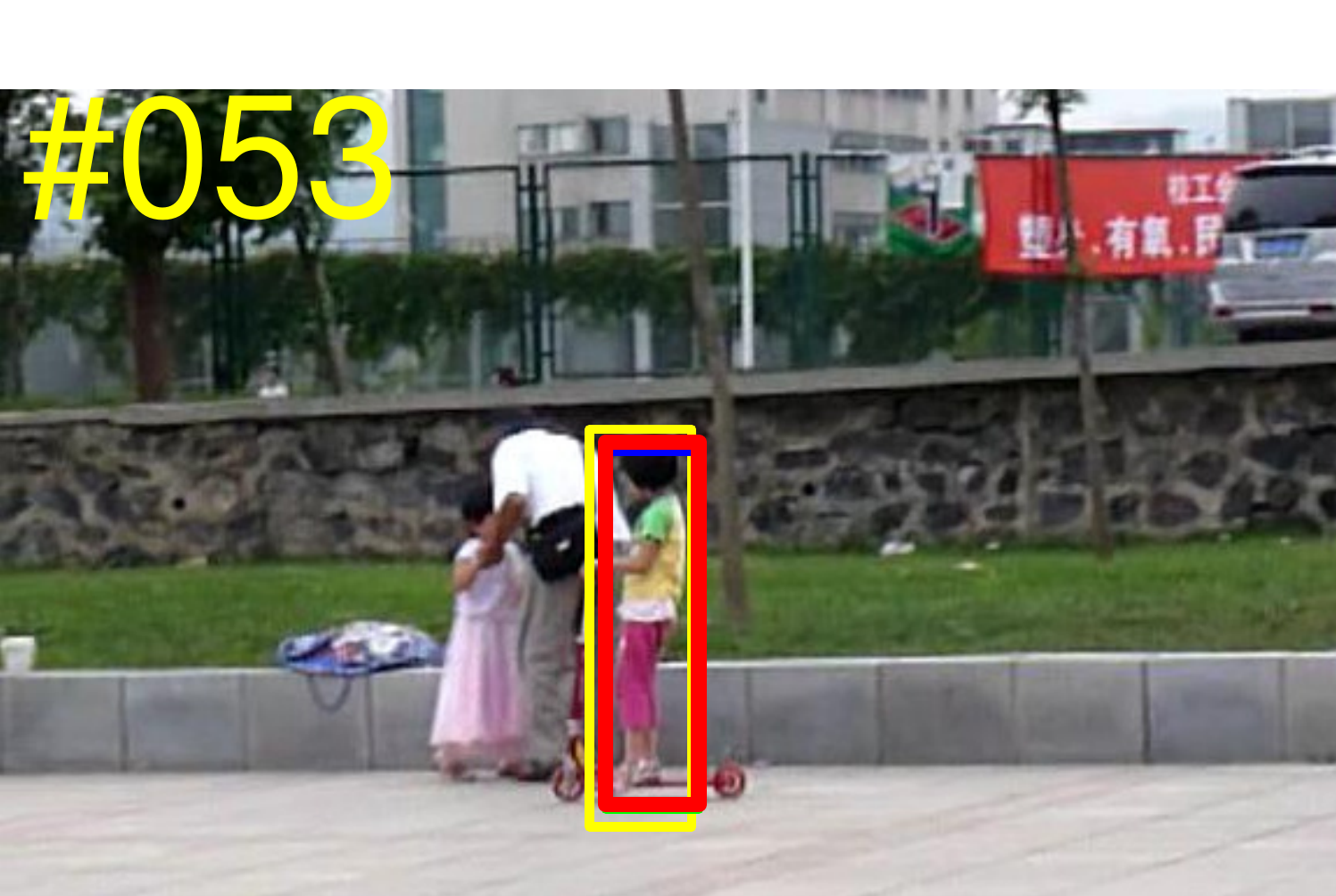}%
	\includegraphics*[trim = 0 100 50 0, width = \wid]{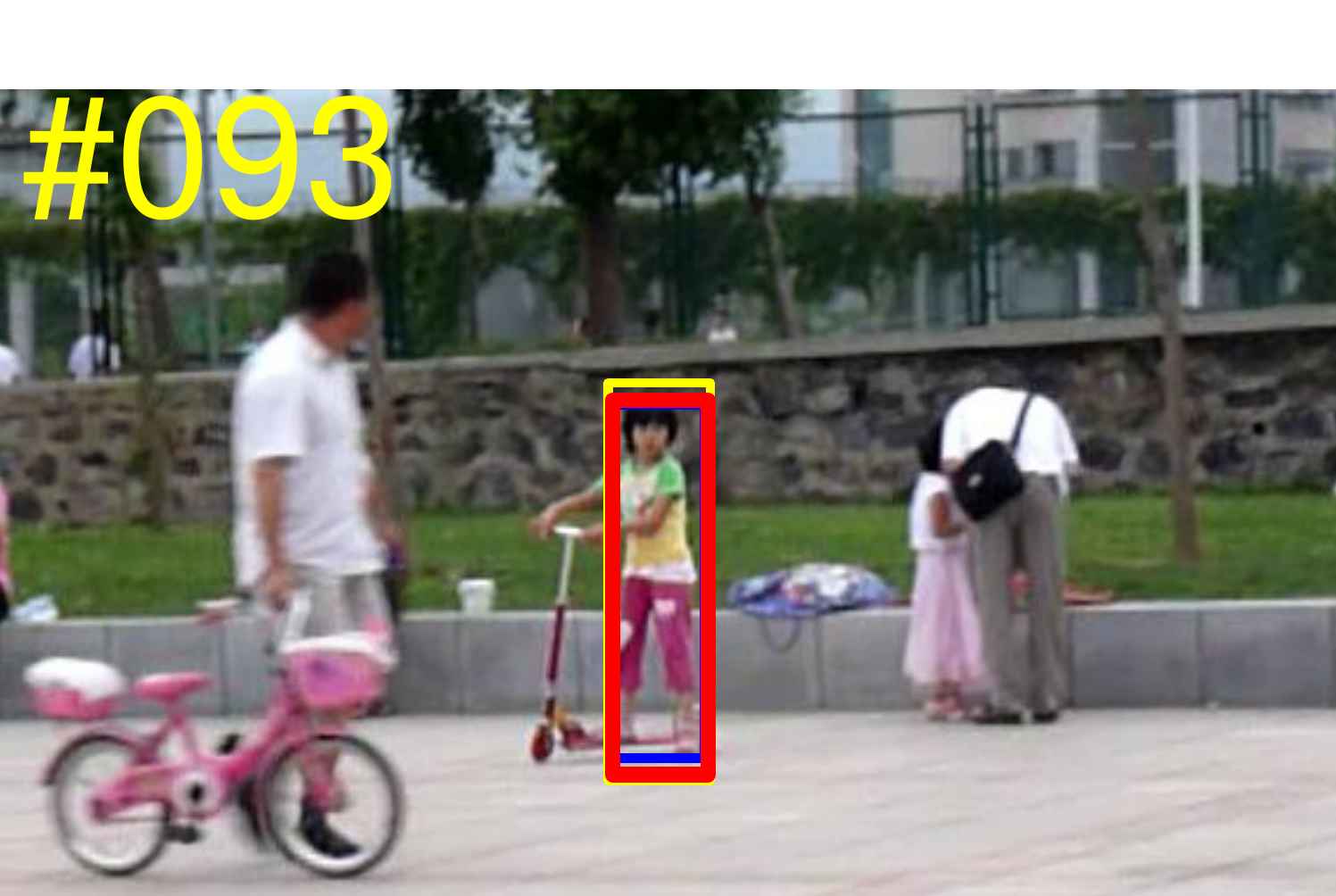}%
	\includegraphics*[trim = 0 100 50 0, width = \wid]{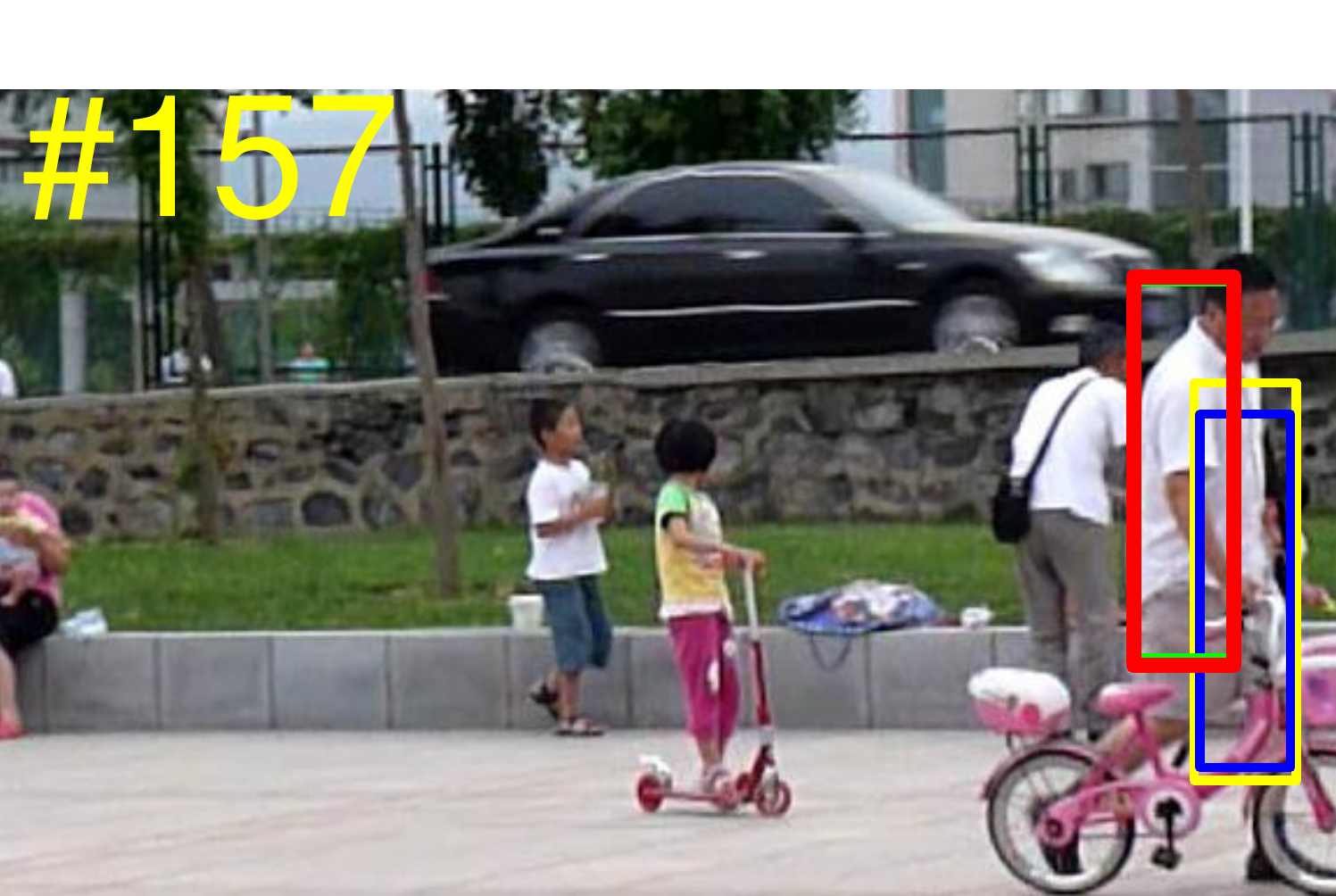}%
	\includegraphics*[trim = 0 100 50 0, width = \wid]{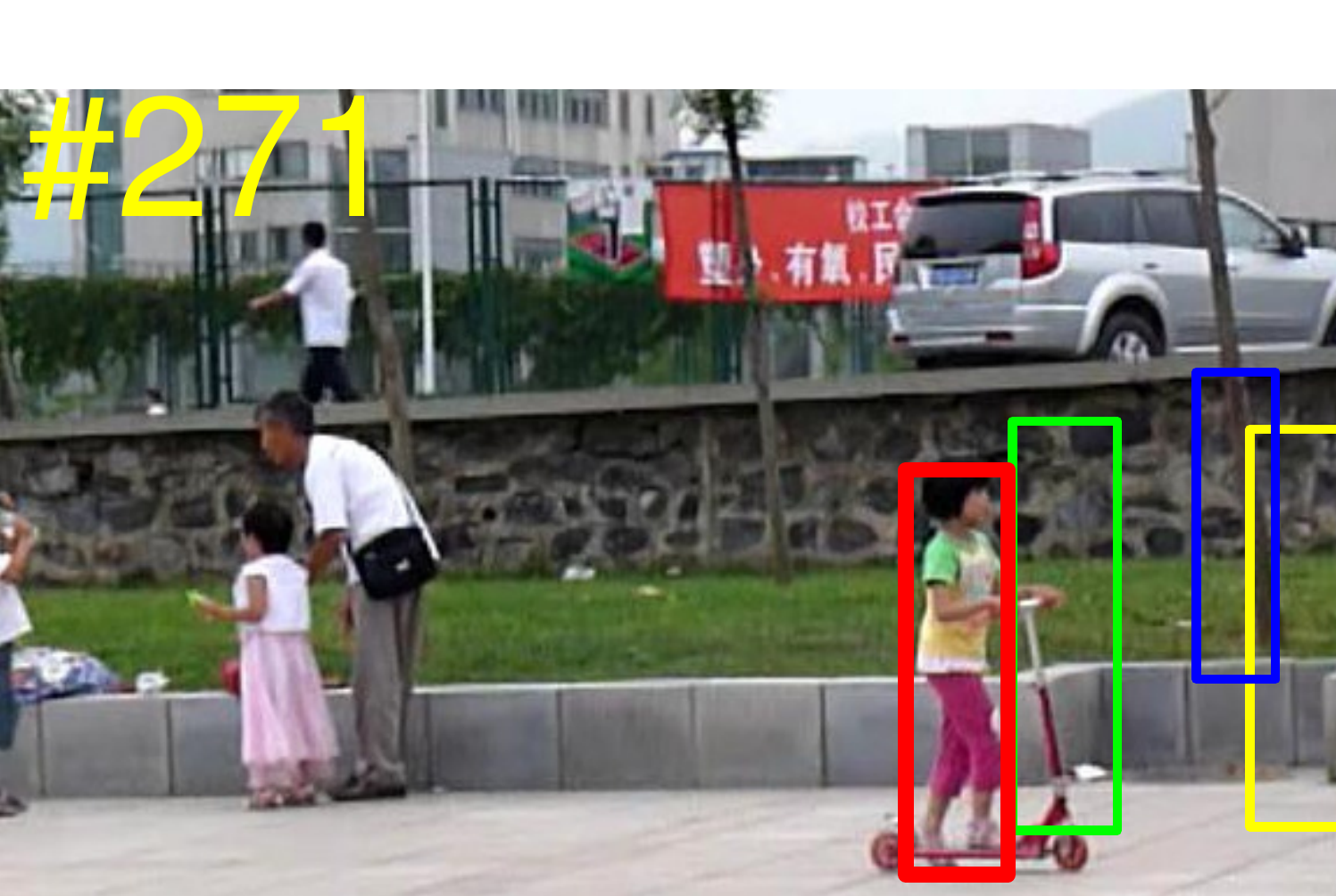}
	\includegraphics*[trim = 2 2 2 5, width = 0.4\textwidth]{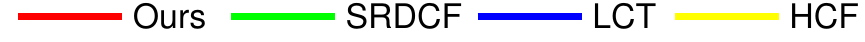}\vspace{-1.5mm}
	\caption{A qualitative comparison of our approach with state-of-the-art methods on the \emph{Box} (top row) and \emph{Girl} (bottom row) videos. Our approach accurately re-detects the target in the \emph{Girl} video due to a decontaminated training set (last frame).}
	\label{fig:qualitative}
	\vspace{-1mm}
\end{figure}

\begin{figure}[!t]
	\centering
	\newcommand{\wid}{0.24\textwidth}
	\includegraphics[width = \wid]{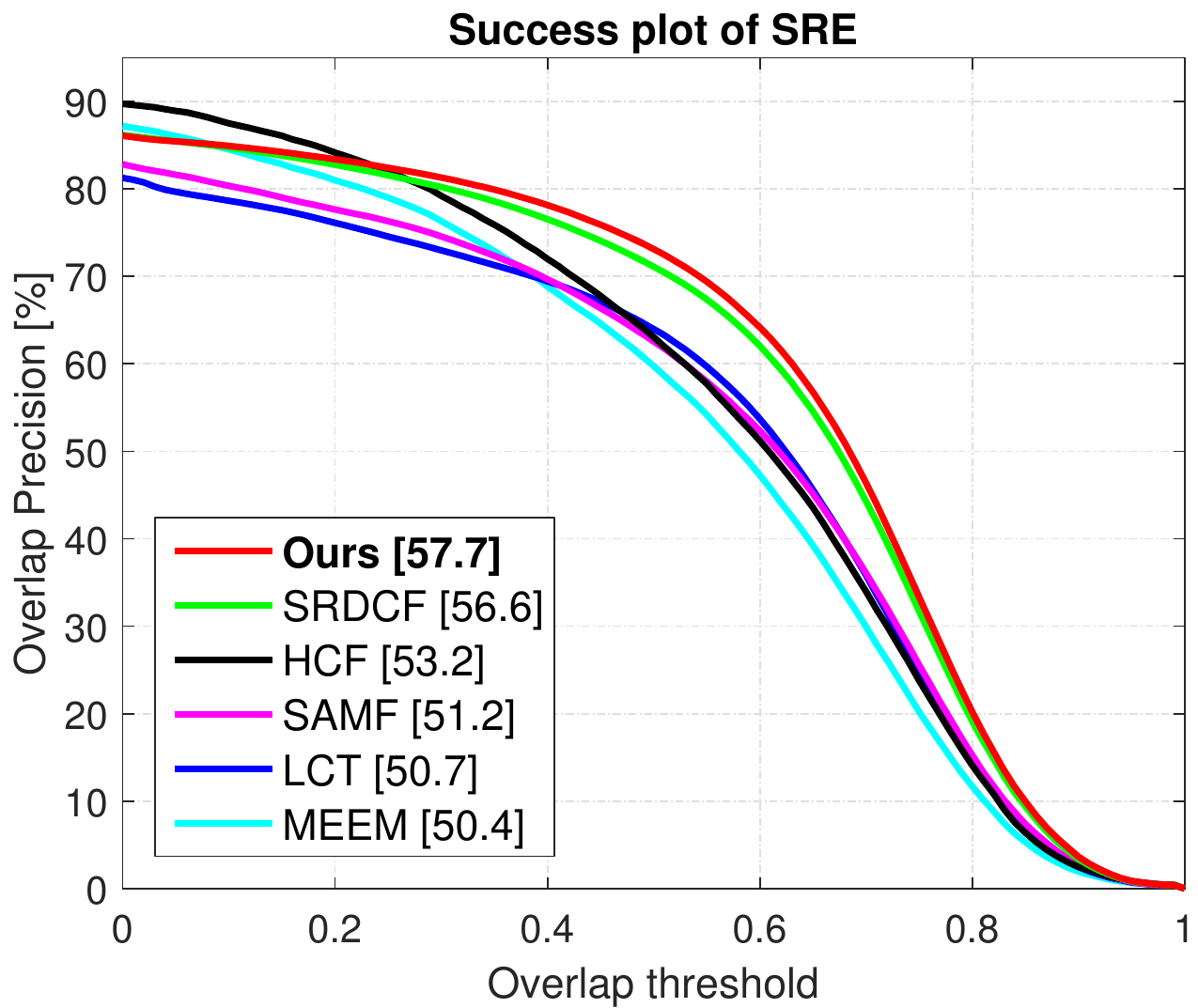}%
	\includegraphics[width = \wid]{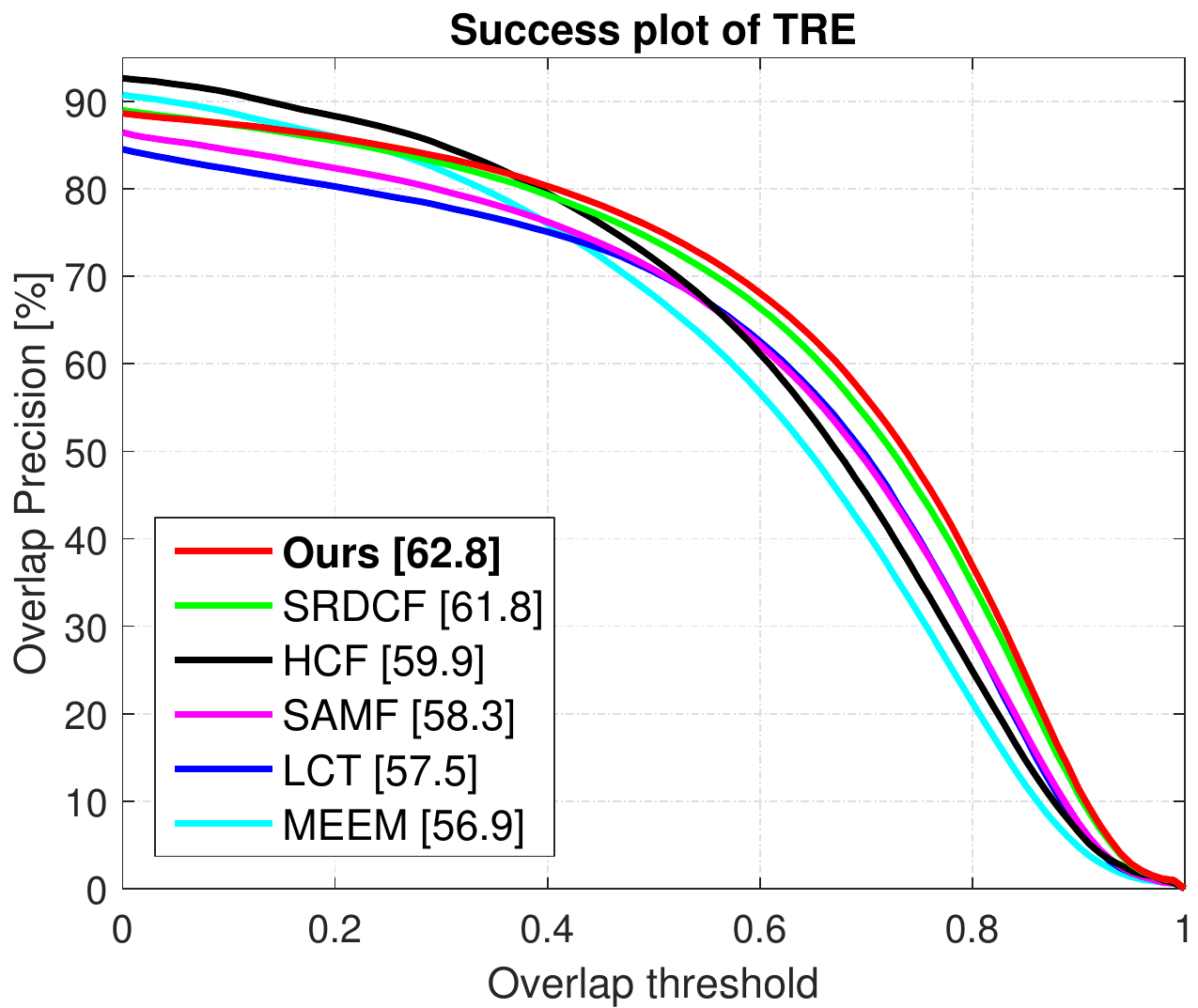}\vspace{-1mm}
	\caption{Robustness to initialization comparison on the OTB-2015 dataset. Success plots are shown for both spatial (SRE) and temporal (TRE) robustness. Our tracker provides consistent improvements in both cases, compared previous approaches.}\vspace{-2mm}
	\label{fig:sota_sretre}
\end{figure}
%% Added by Fahad
\begin{figure*}[!t]
	\centering
	\newcommand{\wid}{0.245\textwidth}
	\newcommand{\name}{figures/sota_OTB100}
	\newcommand{\eval}{OPE}
	\includegraphics[width=\wid]{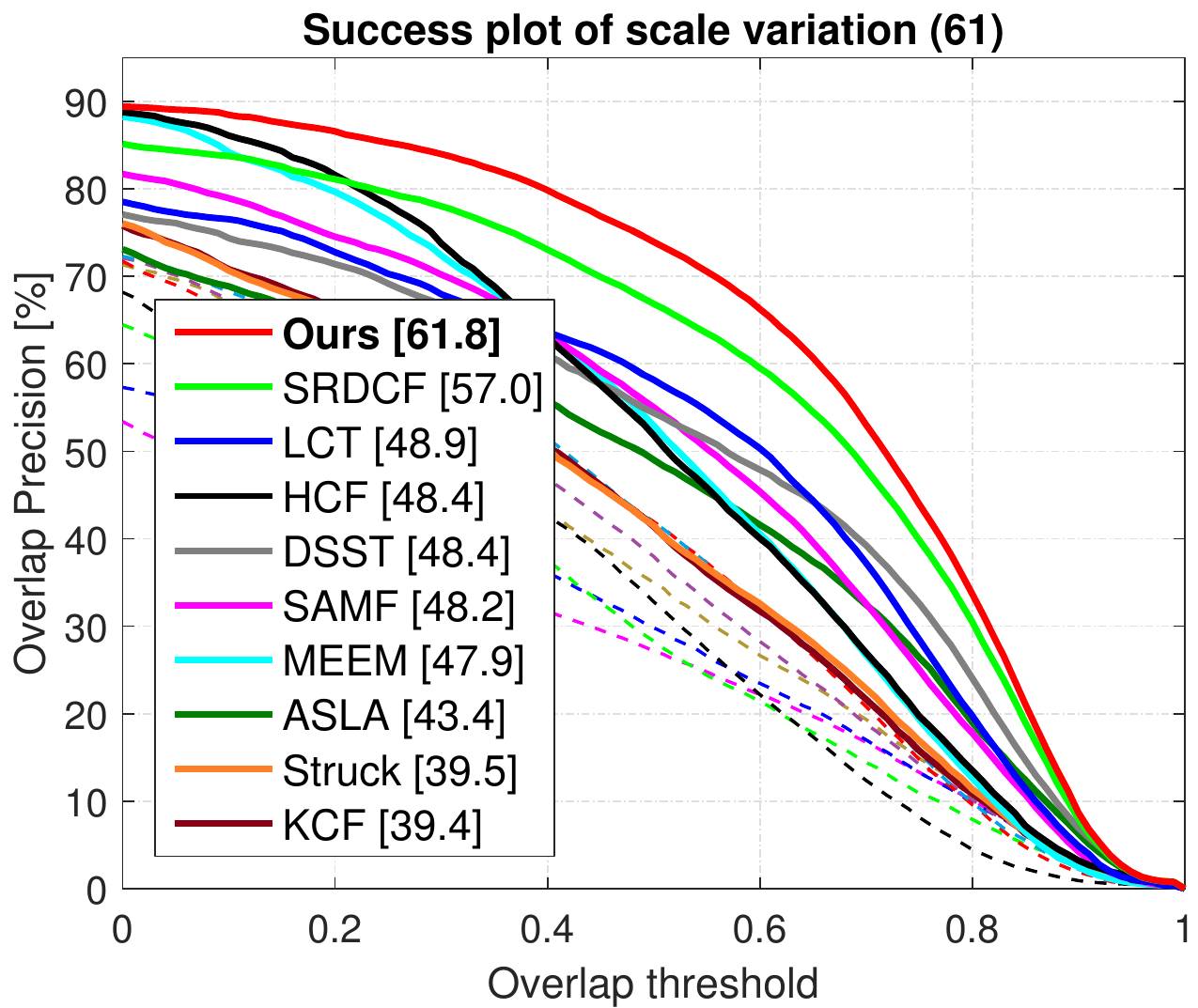}
	\includegraphics[width=\wid]{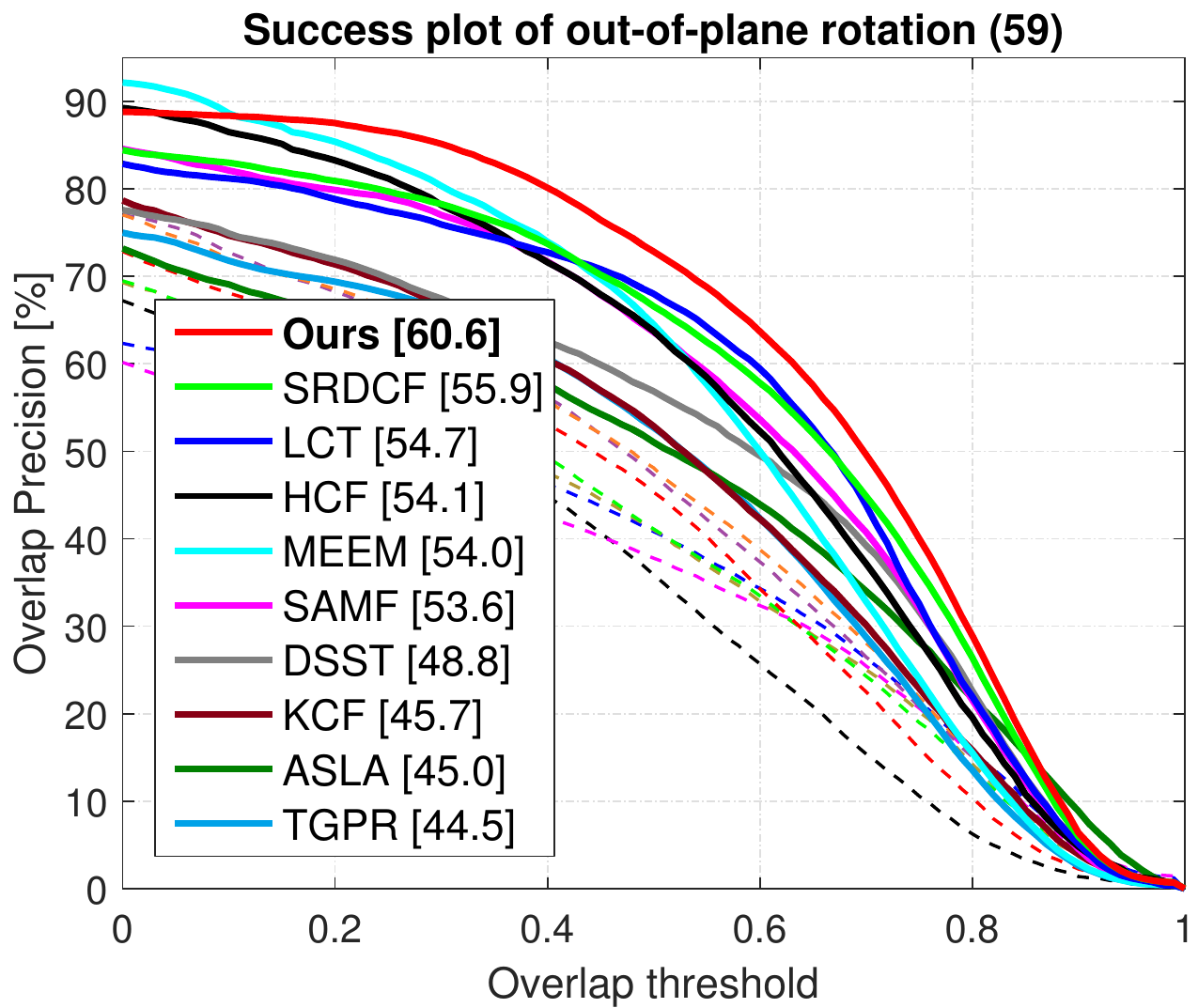}
	\includegraphics[width=\wid]{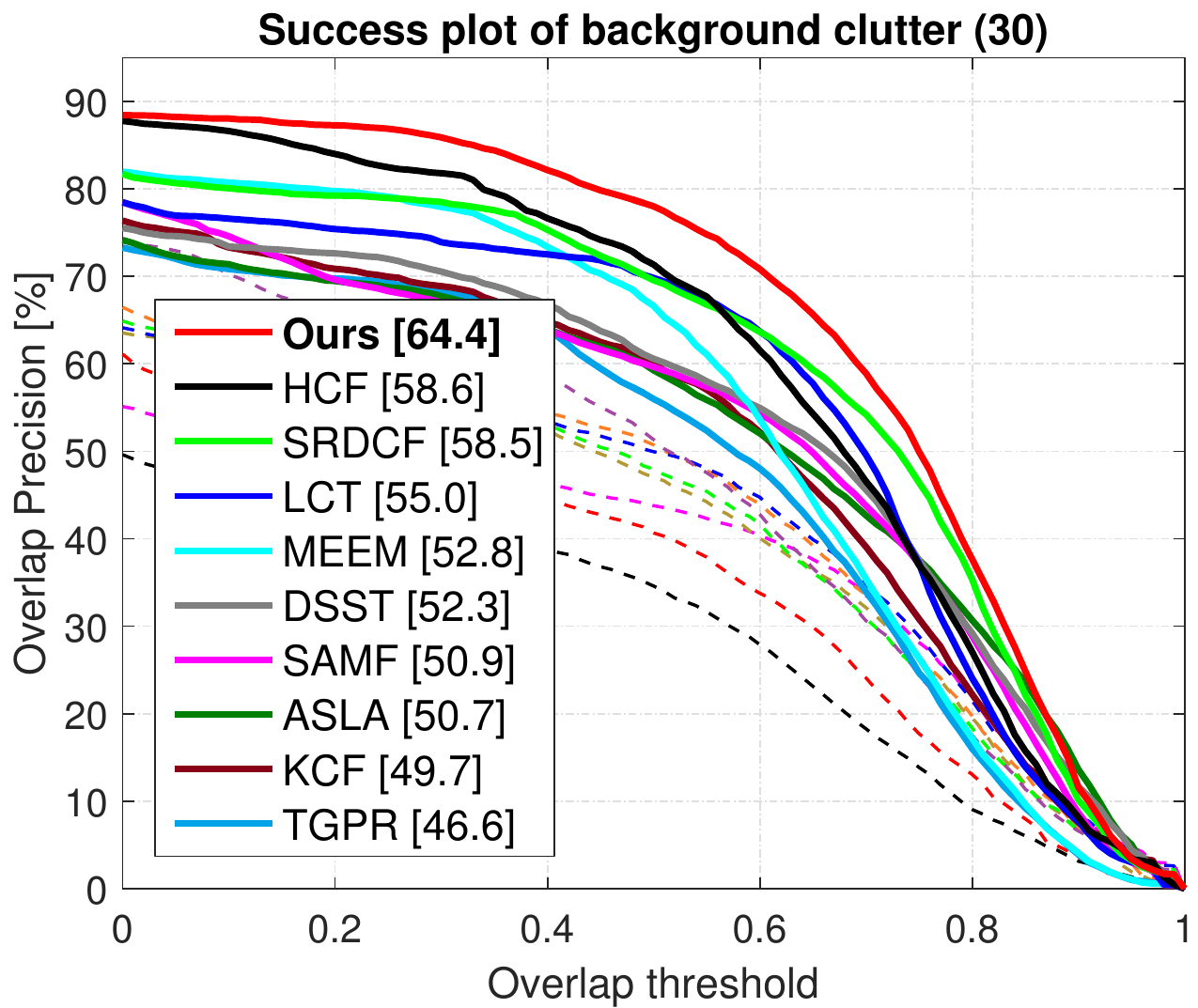}
	\includegraphics[width=\wid]{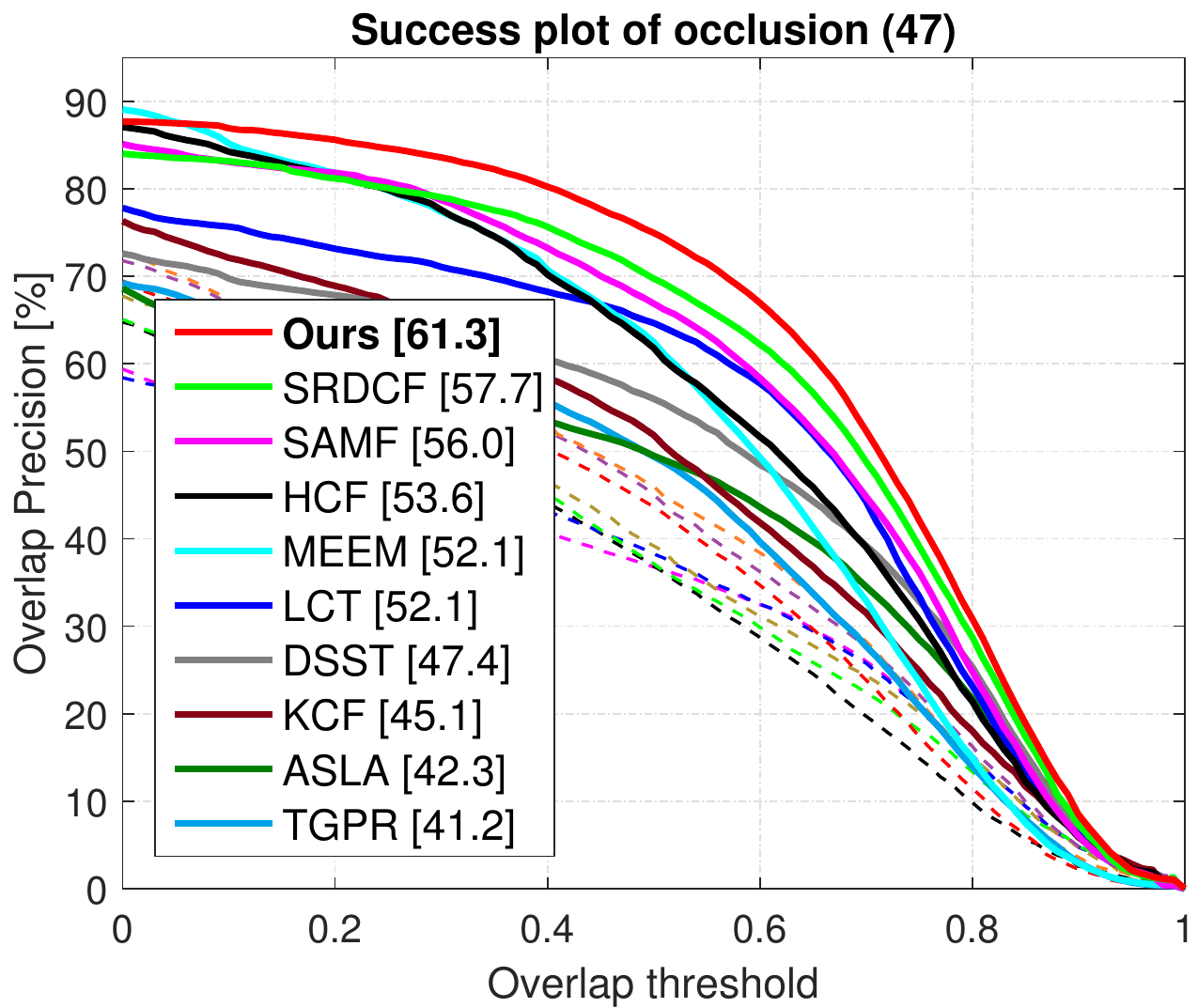}\vspace{-1.5mm}
	\caption{Attribute-based analysis of our approach on the OTB-2015 dataset. Success plots are shown for four attributes where corrupted samples are a common problem. For clarity, we only show the top 10 trackers in the legends. The title of each plot indicates the number of videos labelled with the respective attribute. Our approach provides consistent improvements compared to state-of-the-art methods.}
	\label{fig:attribute}
	\vspace{-3mm}
\end{figure*}

Figure ~\ref{fig:OTB_tpl} contains the success plot, showing the mean OP over the range of overlap thresholds \cite{OTB2015}, on the OTB-2015 dataset.
%The success plot shows the overlap precision (OP) based on the PASCAL overlap criterion over the range of thresholds.
For each tracker, area-under-the-curve (AUC) score is displayed in the legend. Among the compared tracking methods, SRDCF, LCT and HCF provide the best results with AUC scores of $60.5\%$, $56.7\%$ and $56.6\%$ respectively. Our approach achieves the best results with an AUC score of $63.4\%$. Figure~\ref{fig:qualitative} shows a qualitative comparison with state-of-the-art methods on the \emph{Box} and \emph{Girl} videos. Our approach down-weights corrupted training samples, leading to accurate target re-detection (frame 271 in \emph{Girl}).

\subsubsection{Robustness to Initialization}
To evaluate the robustness of our tracker, we follow the protocol proposed by \cite{OTB2015}. The robustness is evaluated using two different initialization strategies: spatial robustness (SRE) and temporal robustness (TRE). The first criteria, SRE, is based on initializing the tracker at different perturbations of the initial ground-truth location.
In case of TRE, the tracker is initialized at 20 different frames with the corresponding ground-truth.
We present the success plots for SRE and TRE, on the OTB-2015, in Figure ~\ref{fig:sota_sretre}. We compare with the top 5 trackers. Our approach achieves robustness in both cases, leading to a consistent performance gain.

\subsubsection{Attribute Based Analysis}
In the OTB-2015, all videos are annotated with 11 different attributes.
Our tracker outperforms previous approaches on all 11 attributes.\footref{exp} Figure~\ref{fig:attribute} shows success plots for four attributes where corrupted samples are commonly included in the training set. In scenarios with challenging scale variations and out-of-plane rotations, inaccurate target estimations often lead to the inclusion of misaligned training samples. Our joint learning approach is capable of reducing or removing the impact of such samples, thereby lowering the risk of drift and tracking failure. In videos with significant background clutter or occlusions, positive training samples are often corrupted by background information. This a common cause for tracking failure in discriminative methods. By re-determining the sample weights in every frame using our joint formulation \eqref{eq:atls}, the effect of corrupted training samples is mitigated by the learning process itself. The effectiveness of our approach is demonstrated by the superior results achieved in the aforementioned scenarios.

\subsection{VOT-2015 Dataset}

In VOT-2015 \cite{VOT2015}, consisting of 60 challenging videos, trackers are evaluated in terms of expected average overlap. This measure is based on empirically estimating the average overlap (as a function of sequence length) and the typical-sequence-length distribution (cutting-off both lopes at a threshold such that the mass is $0.5$). The measure itself is then obtained as the inner product of the two functions. Table~\ref{tab:vot15} shows the average expected overlap (AEO) on VOT-2015 for methods with publicly available implementations.

\begin{table}[!t]
	\centering
	\resizebox{\columnwidth}{!}{%
		\begin{tabular}{l@{~~~}c@{~~}c@{~~}c@{~~}c@{~~}c@{~~}c@{~~}c@{~~}c@{~~}c@{~~}c@{~~}c}
\toprule
 & \textbf{Ours} & SRDCF & MEEM & SAMF & ACT & DSST & KCF & CFLB & Struck & DFT & EDFT \\\midrule
AEO & \textbf{\textcolor{red}{0.299}} & \textit{\textcolor{blue}{0.288}}  & 0.221  & 0.202 & 0.186 & 0.172 & 0.171 & 0.152 & 0.141 & 0.140 & 0.139 \\\bottomrule
%Acc. & 2.45 & 2.28 & 4.67 & 2.92 & 5.03 & 4.55 & 5.45 & 5.45 & 6.78 & 4.62 & 5.22   \\
\end{tabular}

	}
	\vspace{-1.5mm}
	\caption{Comparison with state-of-the-art, based on expected average overlap (EAO), on the VOT-2015 dataset. Our approach provides improved performance compared to the best existing tracker.}
	\label{tab:vot15}
	\vspace{-3mm}
\end{table}

\subsection{Temple-Color Dataset}
Finally, we perform experiments on the Temple-Color dataset with 128 videos. 
A comparison with state-of-the-art trackers is shown in Table~\ref{tab:OTB100} (second row). Among the compared methods, both MEEM and SRDCF obtains a mean OP of $62.2\%$. Our approach improves the state-of-the-art on this dataset with a mean OP of $65.8\%$. Figure~\ref{fig:OTB_tpl} shows the success plot over all the 128 videos in the Temple-Color dataset. MEEM and SRDCF provide AUC scores of $50.6\%$ and $51.6\%$ respectively. Our tracker outperforms state-of-the-art approaches an AUC score of $54.1\%$.

\section{Conclusions}
We propose a unified learning formulation to counter the problem of corrupted training samples in the tracking-by-detection paradigm. Our approach efficiently down-weights the impact of corrupted training samples, while up-weighting accurate samples. The proposed approach is generic and can be integrated into other discriminative tracking frameworks. Experiments demonstrate that our approach achieves state-of-the-art tracking performance.

\noindent\textbf{Acknowledgments}:
This work has been supported by SSF (CUAS), VR (EMC${}^2$ and ELLIIT), the Wallenberg Autonomous Systems Program, the NSC and Nvidia.

%%%%%%%%% BODY TEXT

{\small
\bibliographystyle{ieee}
\bibliography{references}
}

\end{document}